\documentclass[%
 aip,
 amsmath,amssymb,
 reprint,%
]{revtex4-1}
\usepackage{graphicx}
\usepackage{dcolumn}
\usepackage{bm}

\usepackage[utf8]{inputenc}
\usepackage[T1]{fontenc}
\usepackage{mathptmx}
\usepackage{etoolbox}
\usepackage{subcaption}
\usepackage{rotating}
\usepackage{placeins}
\usepackage{afterpage}
\usepackage{placeins}
\usepackage{xcolor}

\begin{document}
\makeatletter
\def\@email#1#2{%
 \endgroup
 \patchcmd{\titleblock@produce}
  {\frontmatter@RRAPformat}
  {\frontmatter@RRAPformat{\produce@RRAP{*#1\href{mailto:#2}{#2}}}\frontmatter@RRAPformat}
  {}{}
}%
\makeatother

\preprint{AIP/123-QED}

\title[DrivAer Transformer]{DrivAer Transformer:\\A high-precision and fast prediction method for vehicle aerodynamic drag coefficient based on the DrivAerNet++ dataset}

\author{Jiaqi He}
\affiliation{Hubei Key Laboratory of Advanced Technology of Automotive Parts, Wuhan University of Technology, Wuhan, China.}

\author{Xiangwen Luo}
\affiliation{Wuhan National Laboratory for Optoelectronic, Huazhong University of Science and Technology, Wuhan, China.}

\author{Yiping Wang}
\email{wangyiping@whut.edu.cn}
\affiliation{Hubei Key Laboratory of Advanced Technology of Automotive Parts, Wuhan University of Technology, Wuhan, China.}

\begin{abstract}
At the current stage, deep learning-based methods have demonstrated excellent capabilities in evaluating aerodynamic performance, significantly reducing the time and cost required for traditional computational fluid dynamics (CFD) simulations. However, when faced with the task of processing extremely complex three-dimensional (3D) vehicle models, the lack of large-scale datasets and training resources, coupled with the inherent diversity and complexity of the geometry of different vehicle models, means that the prediction accuracy and versatility of these networks are still not up to the level required for current production. In view of the remarkable success of Transformer models in the field of natural language processing and their strong potential in the field of image processing, this study innovatively proposes a point cloud learning framework called DrivAer Transformer (DAT). The DAT structure uses the DrivAerNet++ dataset, which contains high-fidelity CFD data of industrial-standard 3D vehicle shapes. Enabling accurate estimation of air drag directly from 3D meshes, thus avoiding the limitations of traditional methods such as 2D image rendering or signed distance fields (SDF). DAT enables fast and accurate drag prediction, driving the evolution of the aerodynamic evaluation process and laying the critical foundation for introducing a data-driven approach to automotive design. The framework is expected to accelerate the vehicle design process and improve development efficiency.
\end{abstract}

\maketitle

\section{Introduction}
With the rapid development of China's electric vehicle industry, OEMs are launching new models at an accelerated rate, and the vehicle development cycle has been shortened dramatically, placing higher demands on efficient and accurate aerodynamic calculations. However, traditional vehicle aerodynamic evaluation relies on a combination of computational fluid dynamics (CFD) and wind tunnel tests, which is a less efficient method that is difficult to meet the needs of vehicle manufacturers to shorten development cycles and control budgets. In addition, CFD requires a high level of experience from engineers, is prone to invalid calculation schemes, and takes a long time for each high-fidelity simulation \cite{aultman2022evaluation, biswas2019development, jiang2021vehicle}. Although the wind tunnel test has high accuracy, it is difficult to cover a large number of design options due to the need to produce a solid scaled model, which is costly and time consuming. In recent years, deep learning-driven methods have been able to rapidly evaluate the aerodynamic performance of different models by utilizing existing datasets, thus significantly speeding up the design process \cite{baque2018geodesic, remelli2020meshsdf, rios2019scalability, rios2021point2ffd, song2023surrogate, trinh20243d, benjamin2025systematic}.

Driven by research in the field of computer vision, multiple deep learning methods for processing 3D models have emerged \cite{atzmon2018point, guo2021pct, li2018pointcnn, qi2017pointnet, qi2017pointnet++, wang2019dynamic}. The ability of object recognition, classification and automatic modeling makes it possible to achieve efficient pneumatic evaluation through deep learning.This type of method can significantly reduce the time and cost required for traditional CFD by building datasets for training with existing model data \cite{remelli2020meshsdf, ribeiro2020deepcfd, rios2019scalability, song2023surrogate}.

However, existing methods still face many challenges when dealing with complex 3D models. For example, large models are limited by training resources and dataset size, while small models have difficulty in balancing accuracy and generalization ability, limiting practical applications\cite{jacob2021deep, tychola2024deep, vinodkumar2024deep, wu2024towards}.

Current research mostly focuses on simplified models such as 2D airfoils \cite{jun2020application, thuerey2020deep, wu2020deep}. For example, Ribeiro's DeepCFD \cite{ribeiro2020deepcfd} uses convolutional neural network (CNN) to quickly simulate 2D laminar velocity and pressure fields. Obiols' CFD Net \cite{obiols2020cfdnet} predicts the flow field characteristics under different boundary conditions with a large amount of model data, which effectively reduces the computational overhead. Although these methods have achieved orders of magnitude speedups in 2D steady state problems, they are still limited in handling complex real-world 3D scenarios. Song \cite{song2023surrogate} predicts the aerodynamic drag coefficients of 3D vehicles with a 2D agent model, and Jacob \cite{jacob2021deep} combines the improved U-Net and the Signed Distance Field (SDF) to predict the aerodynamic properties of 3D vehicle body, which can effectively improve the accuracy, but the geometric enhancement overhead is large and difficult to be used for large-scale data training.

To deal with 3D point clouds, Qi proposes the PointNet architecture \cite{qi2017pointnet}, which can directly deal with unordered point sets, extract global features, and possess rigid-body transformation invariance. Wang proposes DGCNN \cite{wang2019dynamic} to further capture local and global geometric features through dynamic graph structures, and its core EdgeConv operation performs well in classification, segmentation, etc. A follow-up study by Rios has shown that the self-encoder structure can be used to efficiently extract local geometric features in automotive design optimization \cite{rios2021point2ffd}, which improves shape generation and aerodynamic performance optimization.

Elrefaie's recently proposed RegDGCNN\cite{elrefaie2024drivaernet++} improves the DGCNN for regression tasks by introducing a mean-square error loss function for continuous-value prediction, which is suitable for tasks such as aerodynamic parameter regression. Although RegDGCNN performs well in local feature extraction, it still has deficiencies in modeling the overall structural features, especially when facing the new car models that are not seen in the training set, the generalization ability decreases significantly \cite{elrefaie2024drivaernet++}, and the prediction accuracy is not as good as that of CFDs. This exposes the shortcomings of the current model in dealing with the geometry of the complex shapes, which does not make sufficient use of the geometric information. Information is not fully utilized, and new methods are urgently needed to improve it.

To address the above problems, this paper proposes an innovative approach: combining the point cloud attention mechanism (CDA) with a correlation estimation module (CDE), aiming to model the intrinsic geometric information in disordered point cloud structures more effectively. The method implements a complete system capable of accurately predicting aerodynamic drag coefficients based on a 3D STL vehicle model with an accuracy close to that of CFD, while significantly reducing computational resource consumption. After optimization, the system can be adapted to a wide range of vehicle models, which further improves the generalization ability and accuracy of the algorithm.

\section{DrivAerNet++ Dataset}

DrivAerNet++ is the most comprehensive large-scale multimodal dataset for data-driven automotive aerodynamic analysis \cite{elrefaie2024drivaernet++}. It contains 8,000 high-fidelity, industry-standard vehicle models, each with up to 500,000 mesh faces, full 3D flow fields, and precise aerodynamic coefficients. With a 333\% increase in data volume over previous datasets, it is the only open-source database that includes detailed wheel and chassis geometries, making it ideal for advanced aerodynamic modeling.

Compared to traditional datasets like ShapeNet \cite{song2023surrogate}, DrivAerNet++ offers significantly higher geometric detail and design diversity. Conventional datasets often use overly simplified vehicle shapes, lacking critical components such as wheels and underbodies, which results in poor simulation accuracy and limited generalization.

\begin{figure}[h]
    \centering
    \includegraphics[width=0.5\textwidth]{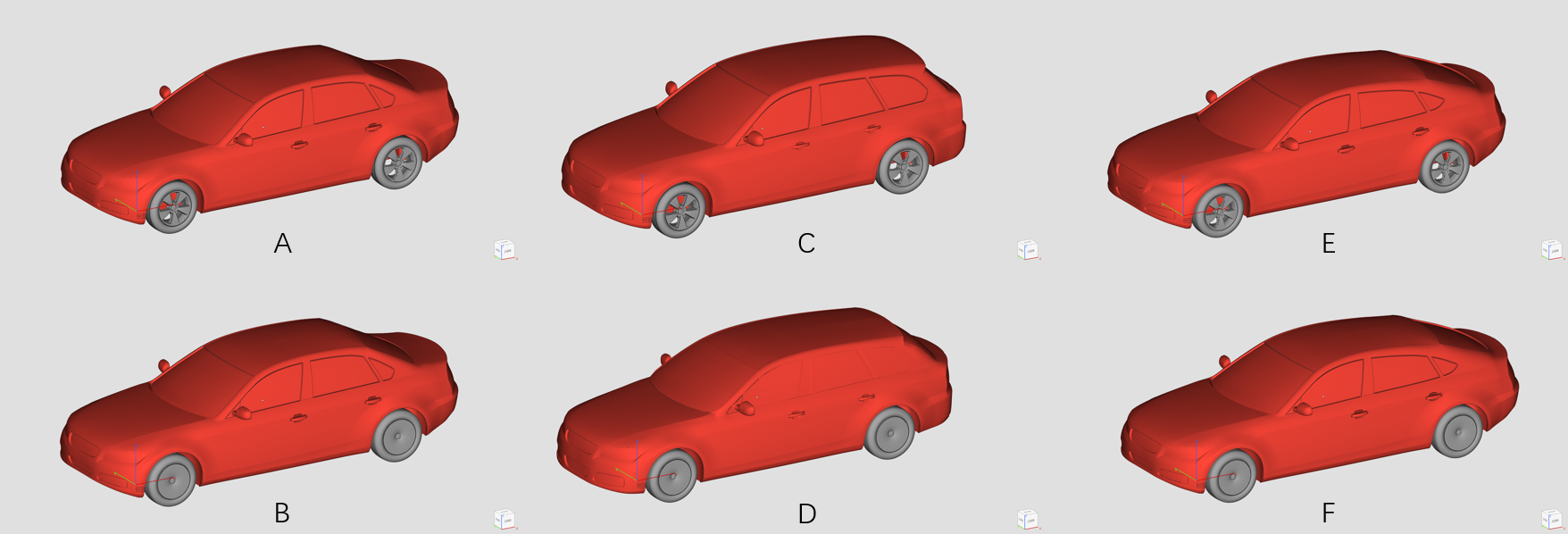}
    \caption{Variation car body types include estateback, fastback, and notchback combined with open and closed wheels.}
    \label{fig:DrivAerNet_image}
\end{figure}

In addition, Fig.~\ref{fig:DrivAerNet_image} presents several example vehicle models from the DrivAerNet++ dataset. These models cover a wide range of vehicle types and design styles, showcasing both high fidelity in geometric detail and broad coverage across different car categories.

\subsection{CFD Simulation Setup}
The study utilizes the DrivAerNet dataset generated through computational fluid dynamics (CFD) simulations implemented in the open-source OpenFOAM\textsuperscript{\textregistered} framework \cite{elrefaie2024drivaernet++, greenshields2022openfoam}. A pressure-velocity coupled steady-state solver (\texttt{$simpleFoam$}) optimized for incompressible turbulent flow analysis was employed. The computational domain, constructed with 1:1 scale representation of the DrivAer fastback vehicle configuration, features dimensions of $5.0L\times3.5L\times2.5L$ (where $L$ denotes vehicle length), incorporating symmetry boundary conditions along the $y$-normal plane to enhance computational efficiency.

\subsection{Turbulence Modeling}
Turbulence closure was achieved through Menter's modified $k$-$\omega$ SST model, which employs blending functions to transition smoothly between near-wall $k$-$\omega$ formulation and free-stream $k$-$\epsilon$ methodology \cite{menter2003ten}. This approach enhances flow separation prediction accuracy under adverse pressure gradients. Governing equations encompass turbulent kinetic energy transport, specific dissipation rate, and eddy viscosity models. Boundary conditions include: uniform inflow velocity (30 m/s, corresponding to Reynolds number $\text{Re}= 9.39\times10^{6}$ based on vehicle length), zero-gradient velocity with fixed static pressure at outlet (configured with anti-backflow treatment), no-slip conditions on vehicle surfaces, rotating wall boundaries for wheels, and slip conditions for domain top/sides. Near-wall resolution was maintained using Spalding-law-based \texttt{$nutUSpaldingWallFunction$} to ensure $Y^+ \approx 1$ \cite{launder1983numerical}.

\begin{figure*}[t]
    \centering
    \begin{subfigure}{0.4\textwidth}
        \centering
        \includegraphics[width=\textwidth]{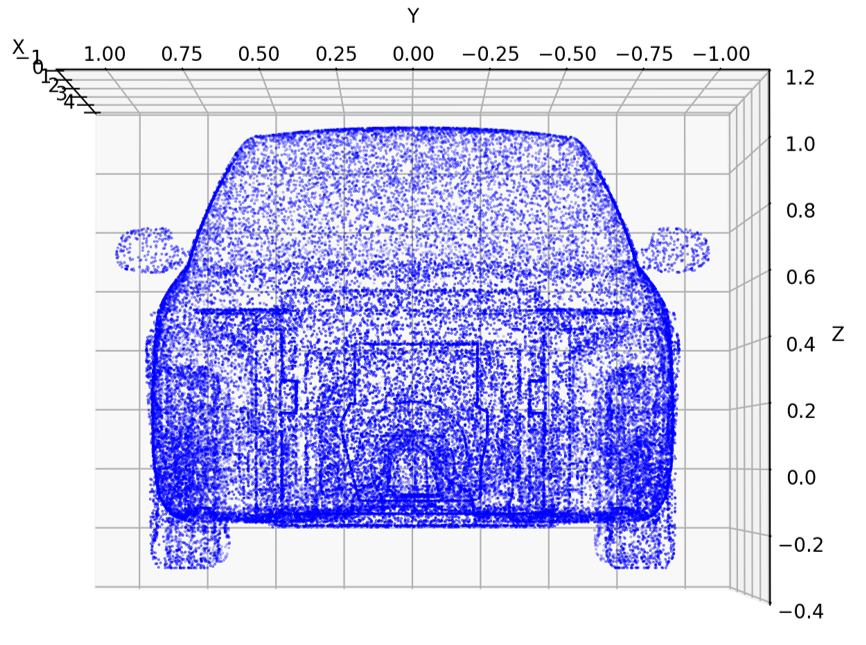}
        \caption{Raw model (with reduced point cloud size)}
    \end{subfigure}
    \hfill
    \begin{subfigure}{0.4\textwidth}
        \centering
        \includegraphics[width=\textwidth]{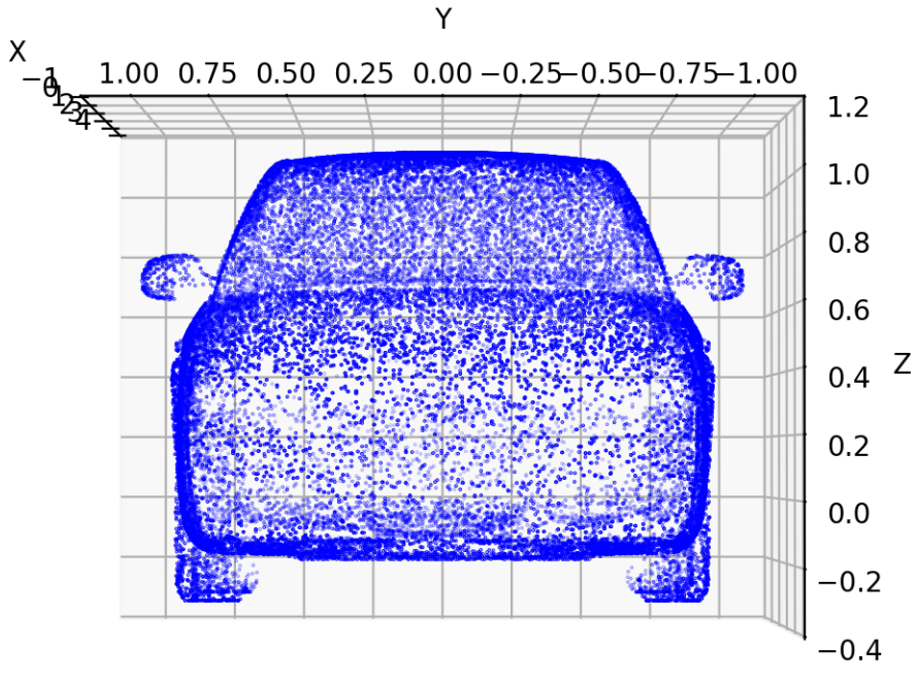}
        \caption{Model after contour extraction}
    \end{subfigure}
    \caption{DrivAer model along the yz-plane}
    \label{fig:yz}
\end{figure*}

\begin{figure*}[t]
    \centering
    \begin{subfigure}{0.4\textwidth}
        \centering
        \includegraphics[width=\textwidth]{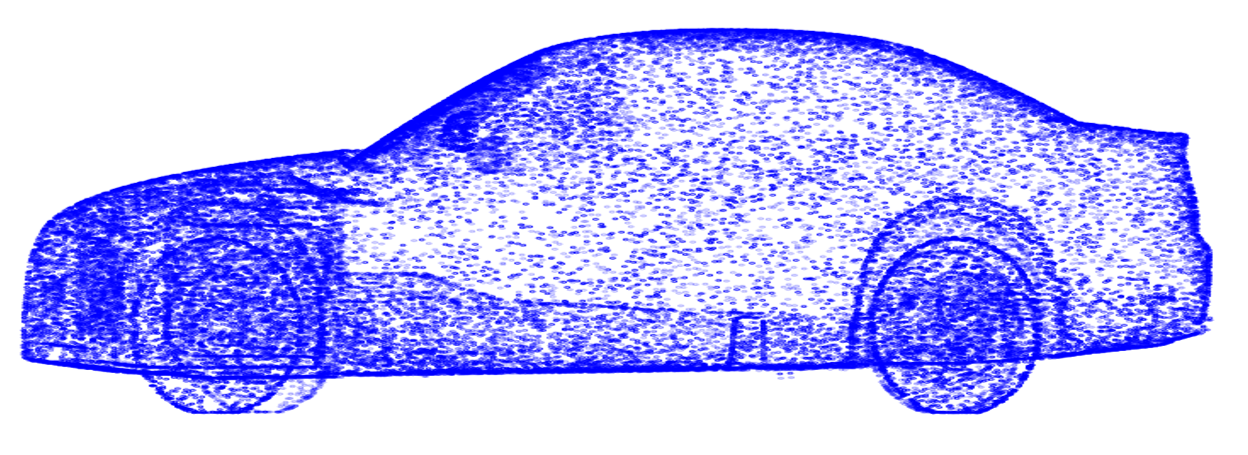}
        \caption{Raw model (with reduced point cloud size)}
    \end{subfigure}
    \hfill
    \begin{subfigure}{0.4\textwidth}
        \centering
        \includegraphics[width=\textwidth]{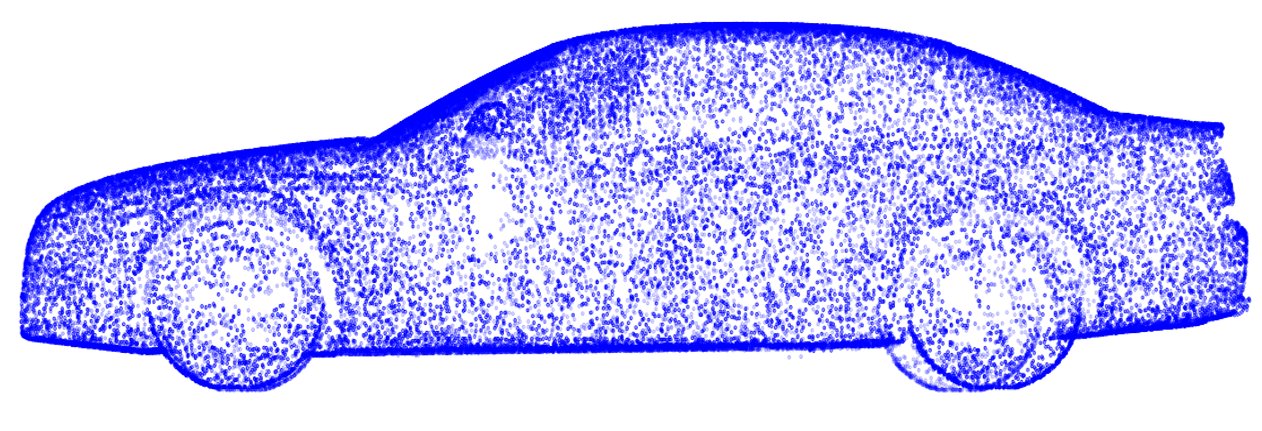}
        \caption{Model after contour extraction}
    \end{subfigure}
    \caption{DrivAer model along the xz-plane}
    \label{fig:xz}
\end{figure*}

\subsection{Grid Generation}
Hex-dominant meshes generated via \texttt{$SnappyHexMesh$} contained 15 prism layers with initial height 0.1 mm and expansion ratio 1.2. Four-tier local refinement zones were implemented:
\begin{itemize}
    \item Vehicle proximity ($\Delta x = 5$ mm)
    \item Wake core ($\Delta x = 10$ mm)
    \item Lateral flow ($\Delta x = 20$ mm)
    \item Far-field background ($\Delta x = 50$ mm)
\end{itemize}
Grid independence was confirmed with $<1\%$ drag coefficient variation between baseline (8M cells, 88 CPUh) and refined (16M cells, 205 CPUh) meshes.

\subsection{Validation Results}
Validation against experimental data for DrivAer fastback configuration showed:
\begin{itemize}
    \item Drag coefficient discrepancies: 2.81\% (coarse) vs 0.81\% (fine)
    \item Surface pressure correlation $R^2 > 0.95$ with wind tunnel measurements
    \item Wake velocity profile RMS error $<3\%$ compared to PIV data
\end{itemize}
confirming numerical model reliability \cite{heft2012experimental, wieser2014experimental}.

The aerodynamic drag coefficient $C_d$ is mathematically defined as:

\begin{equation}
C_d = \frac{2 F_d}{\rho u_\infty^2 A_{\text{ref}}}
\end{equation}

\noindent where $F_d$ denotes the cumulative resistance force acting on the automotive structure, $\rho$ specifies the ambient air mass density, $u_\infty$ characterizes the undisturbed flow velocity, and $A_{\text{ref}}$ indicates the standardized projection area. The resultant force constitutes two distinct physical mechanisms: aerodynamic pressure differential effects (commonly termed shape resistance) and viscous shear forces induced by boundary layer development.

\subsection{Feature Extraction}
In this study, an Alpha Shape-based contour feature extraction algorithm is employed to capture the streamlined characteristics of 3D vehicle models \cite{edelsbrunner1994three}. This approach leverages multi-layer sectional analysis along the vehicle’s longitudinal axis to extract high-density aerodynamic representations. A series of evenly spaced slicing planes are generated along the driving direction of the vehicle. The geometric intersections between these planes and the vehicle body yield cross-sectional contours, details are shown in the Fig.~\ref{fig:yz} and Fig.~\ref{fig:xz}.

\FloatBarrier
\begin{figure*}[t]
    \centering
    \includegraphics[width=\textwidth]{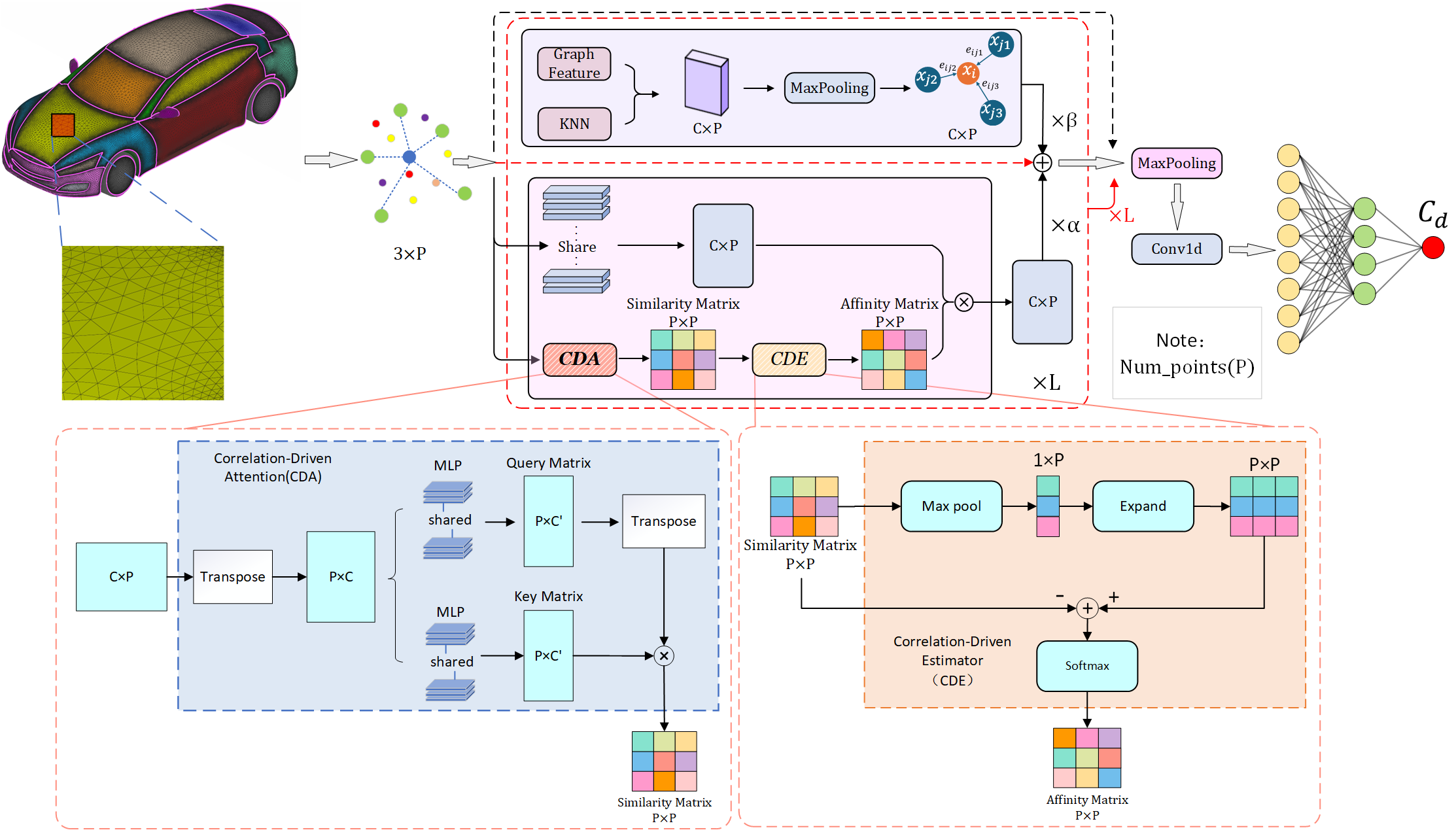}
    \captionof{figure}{\label{fig:dat} A regression framework for predicting drag coefficient. The model processes input mesh points with a 3D automotive mesh model based on a conventional point cloud feature extraction framework \cite{qi2017pointnet, qi2017pointnet++}. The network employs L edge convolution along with a correlative attention fusion module to aggregate point set features and characterize boundary features through a dynamically learned graph network \cite{you2022roland}. The associative attention module utilizes a multi-layer perceptron (MLP) with shared weights to determine edge features, constructing similarity and evaluation matrices. These auto-correlation matrices are dynamically weighted within the dynamic graph space feature extraction module, and the final MLP maps the aggregated 1D feature vectors into drag coefficient.}
\end{figure*}

Edge feature points are automatically identified per cross-section using the Alpha Shape method. An adaptive sampling strategy, sensitive to curvature, increases point density threefold in areas with small curvature radii—such as windshield transitions and rear edges—compared to flatter regions. A total of 1,000 feature points per section form a sparse point cloud preserving key aerodynamic characteristics.

This approach provides flexibility in controlling slices, sampling density, and curvature thresholds. It efficiently retains critical aerodynamic geometry, discards irrelevant internal details, and significantly enhances simulation efficiency without compromising prediction accuracy.

\section{DrivAer Transformer (DAT)}

Recent advancements in geometric deep learning have shown promising potential in solving hydrodynamic problems involving complex geometries~\cite{kashefi2022physics, pfaff2020learning, rios2019scalability, rios2021point2ffd, rios2019efficiency, sanchez2020learning}. In this study, we design a deep learning framework to predict the aerodynamic drag coefficient ($C_d$) of vehicles from 3D point cloud data. The model leverages geometric feature extraction, dynamic graph convolution, and a self-attention mechanism to effectively capture both local and global aerodynamic features. The overall architecture is shown in Fig.~\ref{fig:dat}, with additional visualizations provided in the appendix.

\subsection{A regression prediction framework based on dynamic graph networks}

Aerodynamic drag is highly sensitive to vehicle geometry, necessitating a model that captures both local correlations and global structural patterns in 3D point cloud data. Since point clouds are unordered, conventional convolutional networks aggregate global features using max pooling to maintain permutation invariance. Locally, a graph-based representation enables convolution-like operations across neighboring points, allowing the model to extract flow-related features more precisely.

Building upon the RegDGCNN architecture, this research introduce a Correlation-Driven Attention (CDA) mechanism and a Correlation-Driven Estimator (CDE) to improve structural feature extraction from disordered point clouds. The hybrid framework combines PointNet’s spatial encoding with DGCNN’s local relationship modeling, enabling continuous regression of aerodynamic coefficients. The attention-guided feature bridge enhances both translation invariance and global expressiveness. Multi-layer correlation edge convolutions adaptively update the local graph structure based on dynamic changes in the input features. Given a point cloud $X = \{x_1, \ldots, x_n\} \in \mathbb{R}^F$ and an initial graph $G$, the extracted graph features are denoted $\gamma \in \mathbb{R}^{C \times P}$, where $C$ is the number of channels and $P$ the number of points.

\subsection{Correlation-Driven Attention(CDA)}

While EdgeConv effectively models local geometric features, it lacks the ability to capture global semantic structures necessary for accurate drag prediction. To address this, we propose a global attention mechanism that leverages point cloud feature invariance. This mechanism performs channel-wise dynamic reweighting to highlight critical structural features using learned attention weights, while preserving EdgeConv’s local modeling advantages.

Specifically, the similarity between transposed graph features and the original input features is computed to generate a similarity matrix, thereby enriching the model's understanding of global spatial relationships. This formulation enables the network to emphasize globally important regions while maintaining local detail.

To further enhance expressiveness, a novel activation function parameterized (AconC \cite{ma2021activate}) by learnable variables $p_1$, $p_2$, and $\beta$ is introduced, allowing the model to adapt nonlinearity based on input characteristics. For input $x \in \mathbb{R}^{1 \times C \times 1}$, the activation operates as follows:

\begin{align}
dpx &= (p_1 - p_2 \times x) \\
AconC &= dpx \times \sigma(\beta \times dpx) + p_2 \times x \\
G^q &= AconC(\tau(\text{Conv}(\gamma^T))^T) \\
G^k &= AconC(\tau(\text{Conv}(\gamma^T))) \\
S &= (G^q)^T \times G^k
\end{align}

\noindent where \(S\) represents the similarity matrix, \(\tau\) denotes batch normalization, \(\sigma\) is the Sigmoid activation function, and \(\gamma^T\) represents the transposed input point cloud features. The AconC activation function introduces a switching factor to learn the parameter transition between non-linear (activation) and linear (non-activation) states. This allows the model to more flexibly select the appropriate mapping approach based on varying point cloud features.

\subsection{Correlation-Driven Estimator(CDE)}

On the basis of the similarity matrix $S$, a novel method is proposed to extend local similarity features into a global structural representation that captures category-specific patterns. The intuition is to treat each channel's features as representations of global semantic information. While local similarity reflects geometric structure at a neighborhood level, effective global modeling of unordered point clouds requires aggregation and weighting of local patterns into a coherent global feature.

Based on the similarity matrix $S$, channel-wise importance is recalculated by leveraging inter-channel affinities. A key operation involves applying an $\arg\max$ over $S$ along the channel dimension to extract the most salient similarity values. These maxima highlight the most informative components in each channel, aiding global structure identification.

The features extracted via the $\arg\max$ operation are then normalized using a Softmax function to enhance their influence across the channel dimension. This activation improves feature contrast and guides the model to focus on the most representative components of global geometry.

The above process can be expressed as:

\begin{equation}
A = \sigma\left(\arg\max(S) - S\right)
\end{equation}

\noindent where \(\arg\max\) denotes the operation that extracts the maximum values along the channel dimension from the similarity matrix, obtaining the most critical information. \(\sigma\) represents the Softmax non-linear activation function. 

This channel affinity mechanism enhances the model’s ability to translate local structural similarities into a discriminative global representation. It adaptively emphasizes the channels contributing most to global semantics, thus improving performance in tasks that require higher level geometric understanding particularly in complex scenarios like aerodynamic drag prediction.

\begin{figure*}[htbp]
    \centering
    \begin{minipage}[b]{0.33\textwidth}
        \centering
        \includegraphics[width=\textwidth]{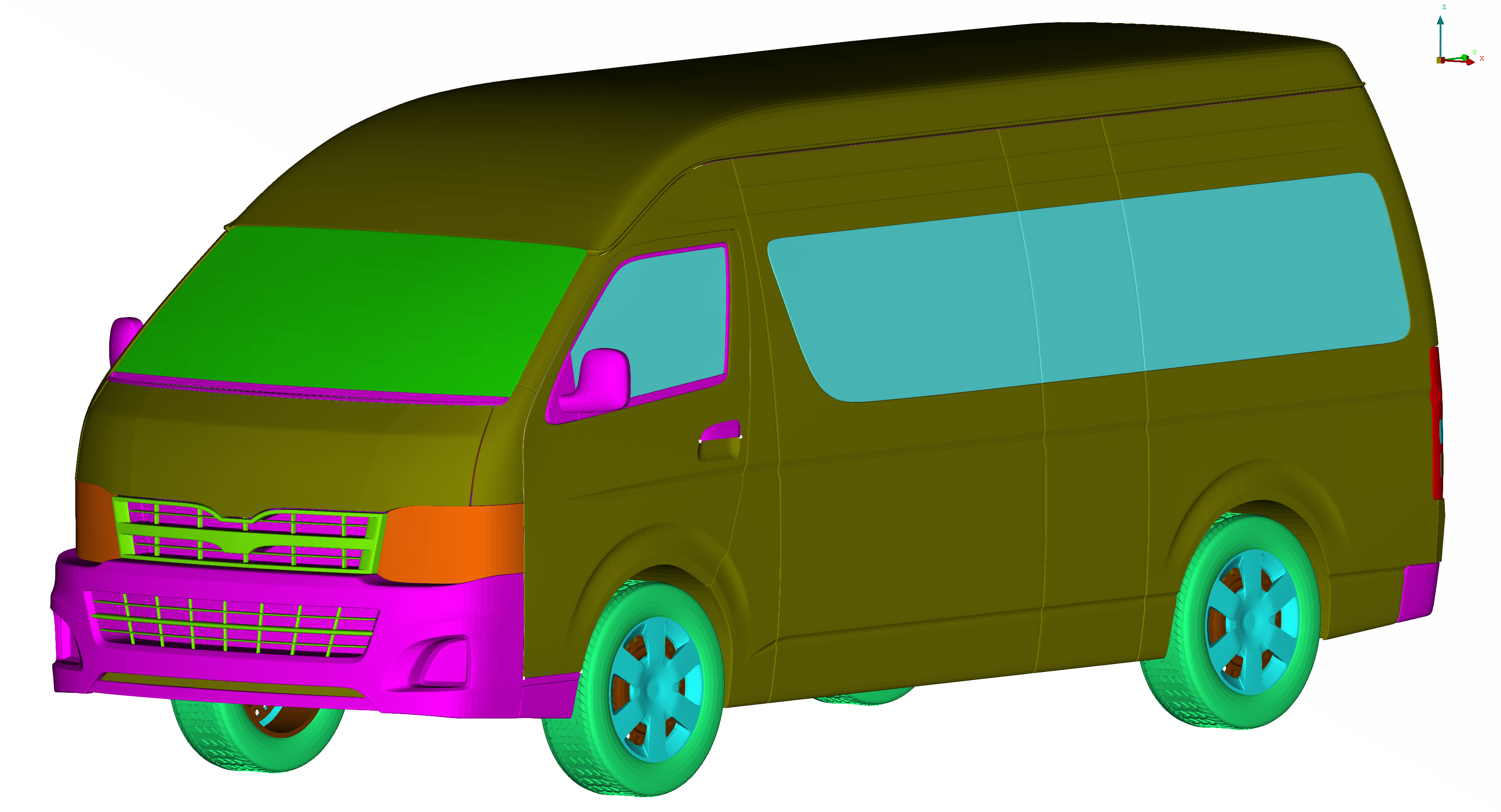}
        \caption*{A. MPV Model}
    \end{minipage}
    \hfill
    \begin{minipage}[b]{0.33\textwidth}
        \centering
        \includegraphics[width=\textwidth]{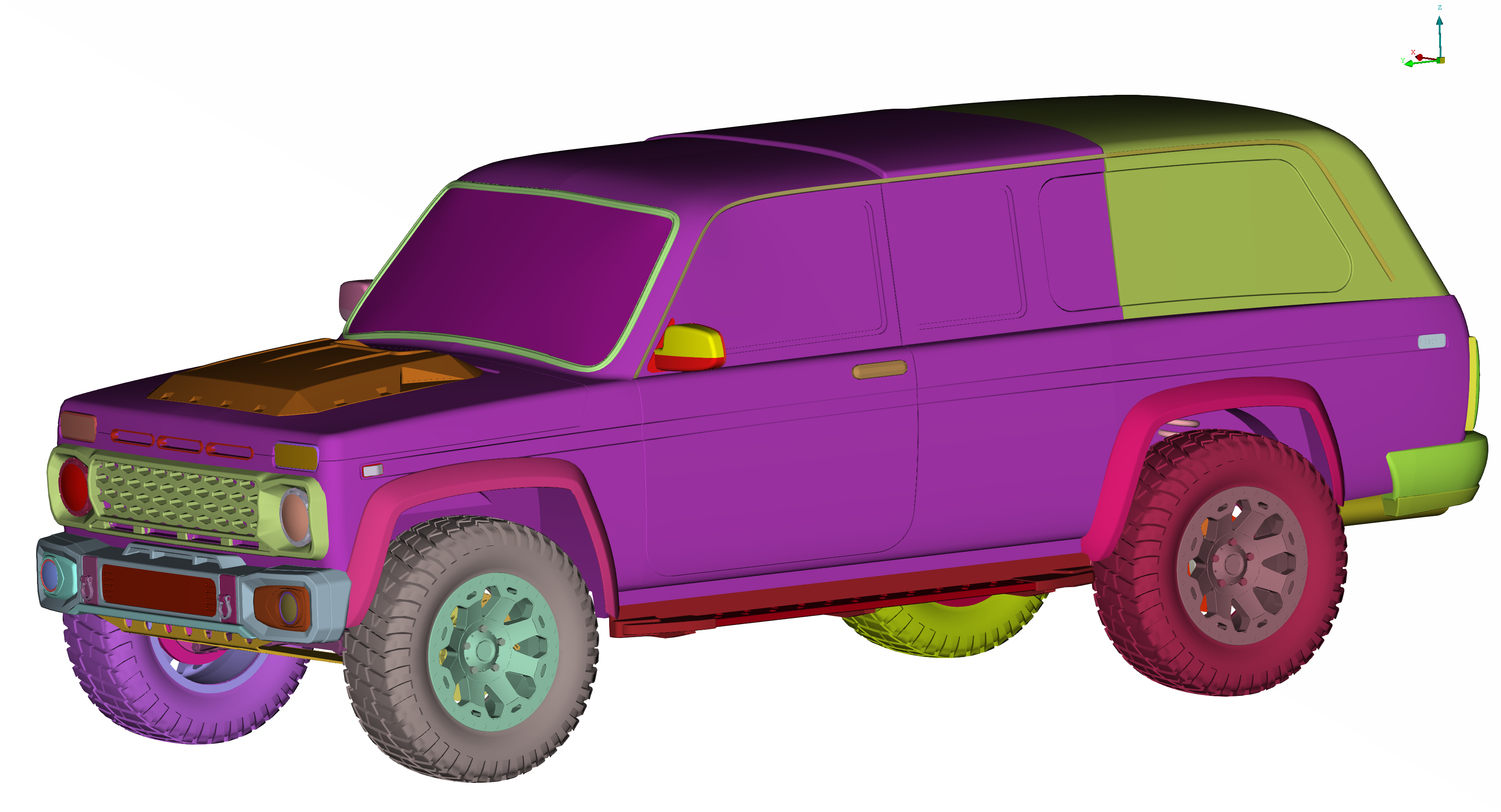}
        \caption*{B. ORV Model}
    \end{minipage}
    \hfill
    \begin{minipage}[b]{0.33\textwidth}
        \centering
        \includegraphics[width=\textwidth]{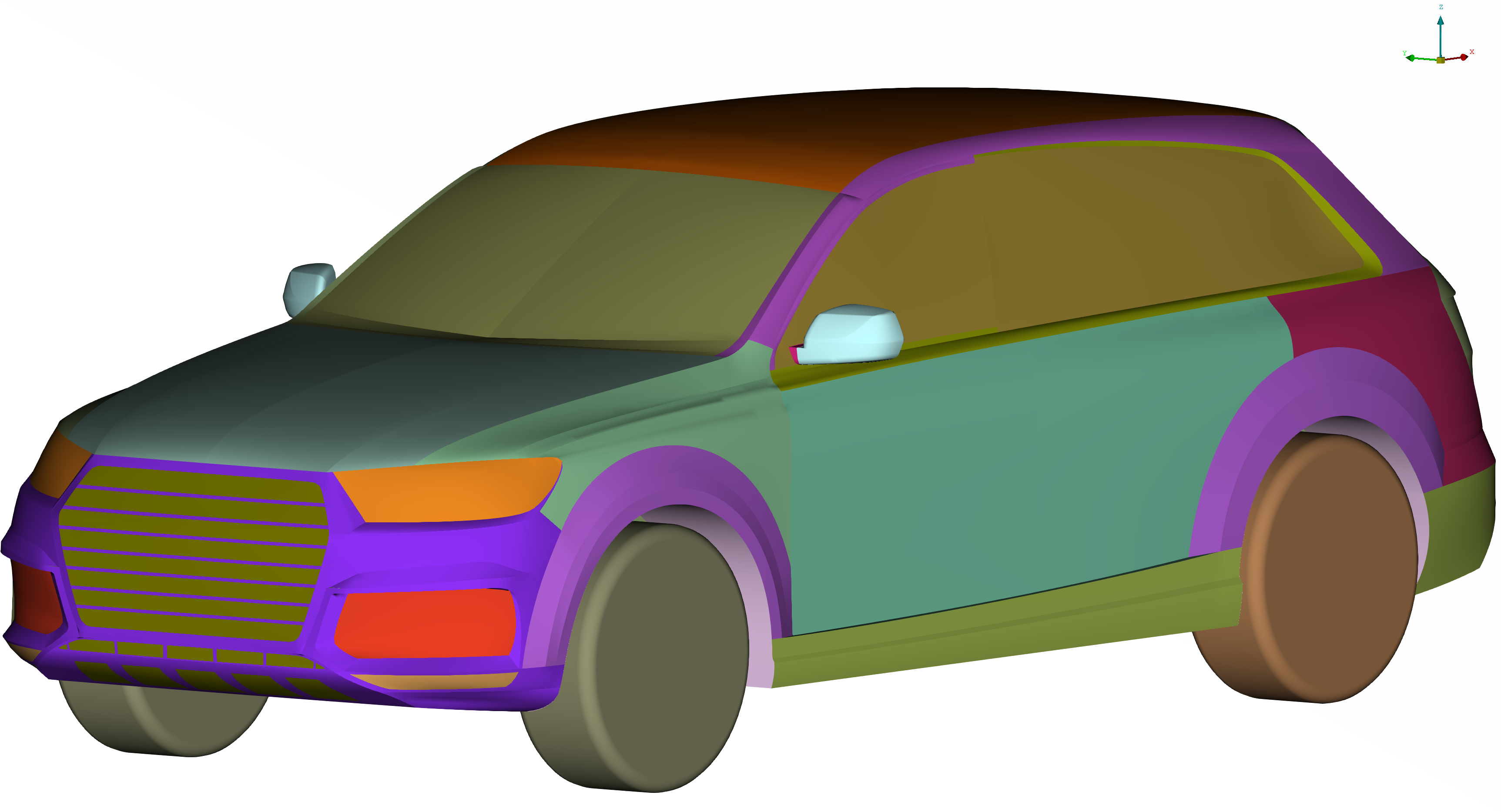}
        \caption*{C. SUV Model}
    \end{minipage}
    \caption{Large-space models sample showcase}
    \label{fig:SUV}
\end{figure*}

\begin{figure*}[htbp]
    \centering
    \begin{minipage}[b]{0.33\textwidth}
        \centering
        \includegraphics[width=\textwidth]{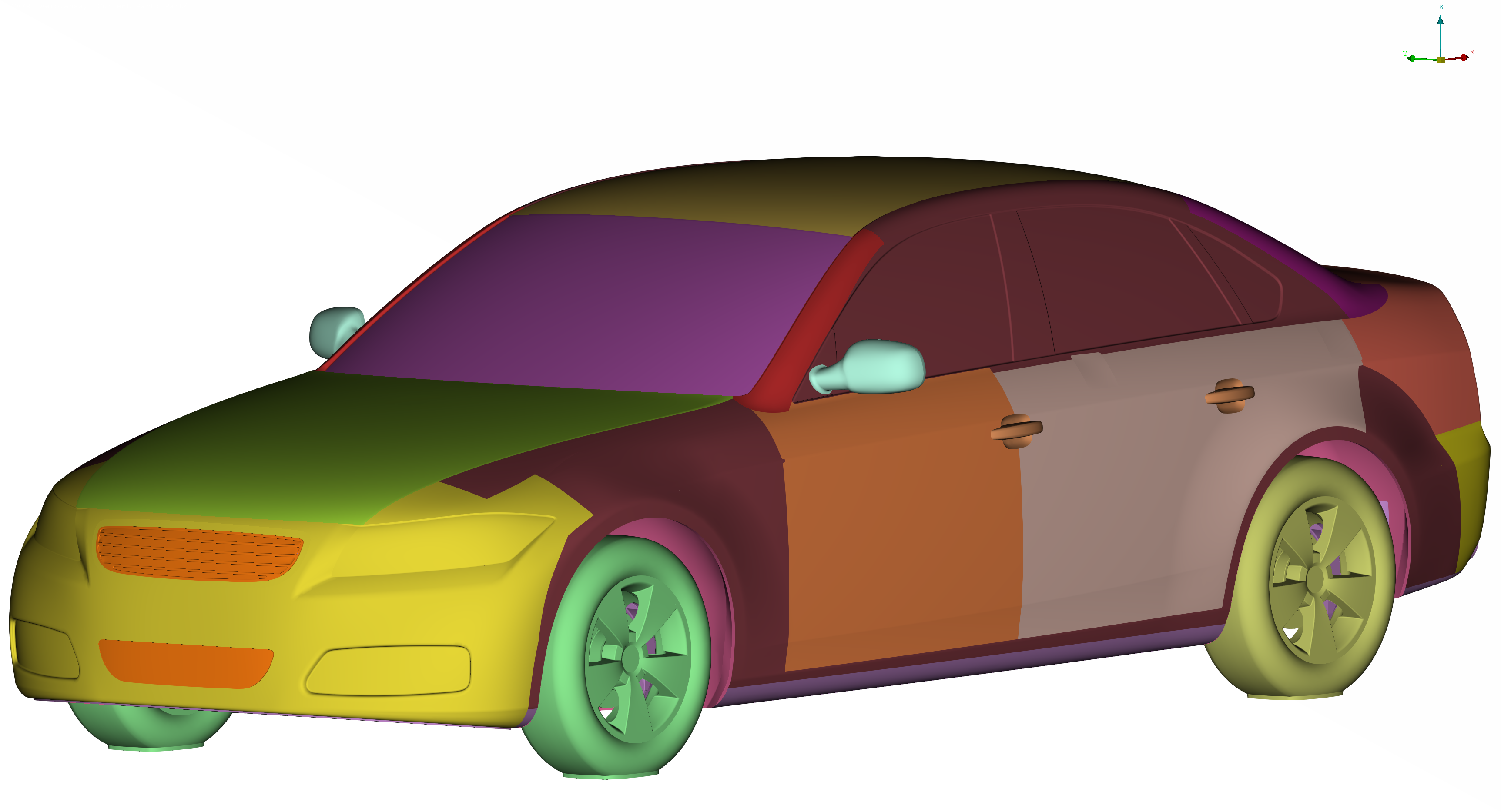}
        \caption*{A. CAERI Aero Model}
    \end{minipage}
    \hfill
    \begin{minipage}[b]{0.33\textwidth}
        \centering
        \includegraphics[width=\textwidth]{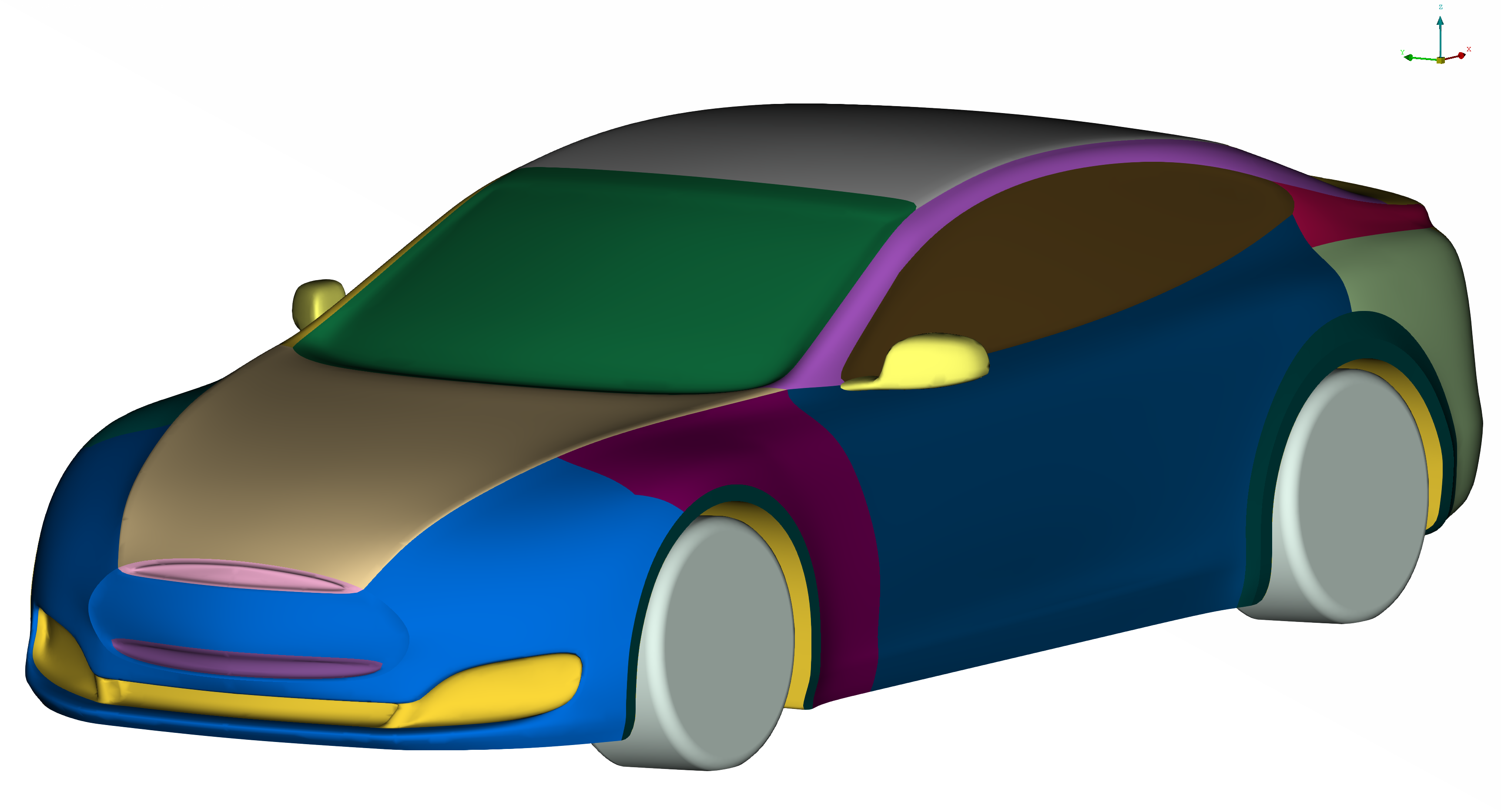}
        \caption*{B. PCV Model}
    \end{minipage}
    \hfill
    \begin{minipage}[b]{0.33\textwidth}
        \centering
        \includegraphics[width=\textwidth]{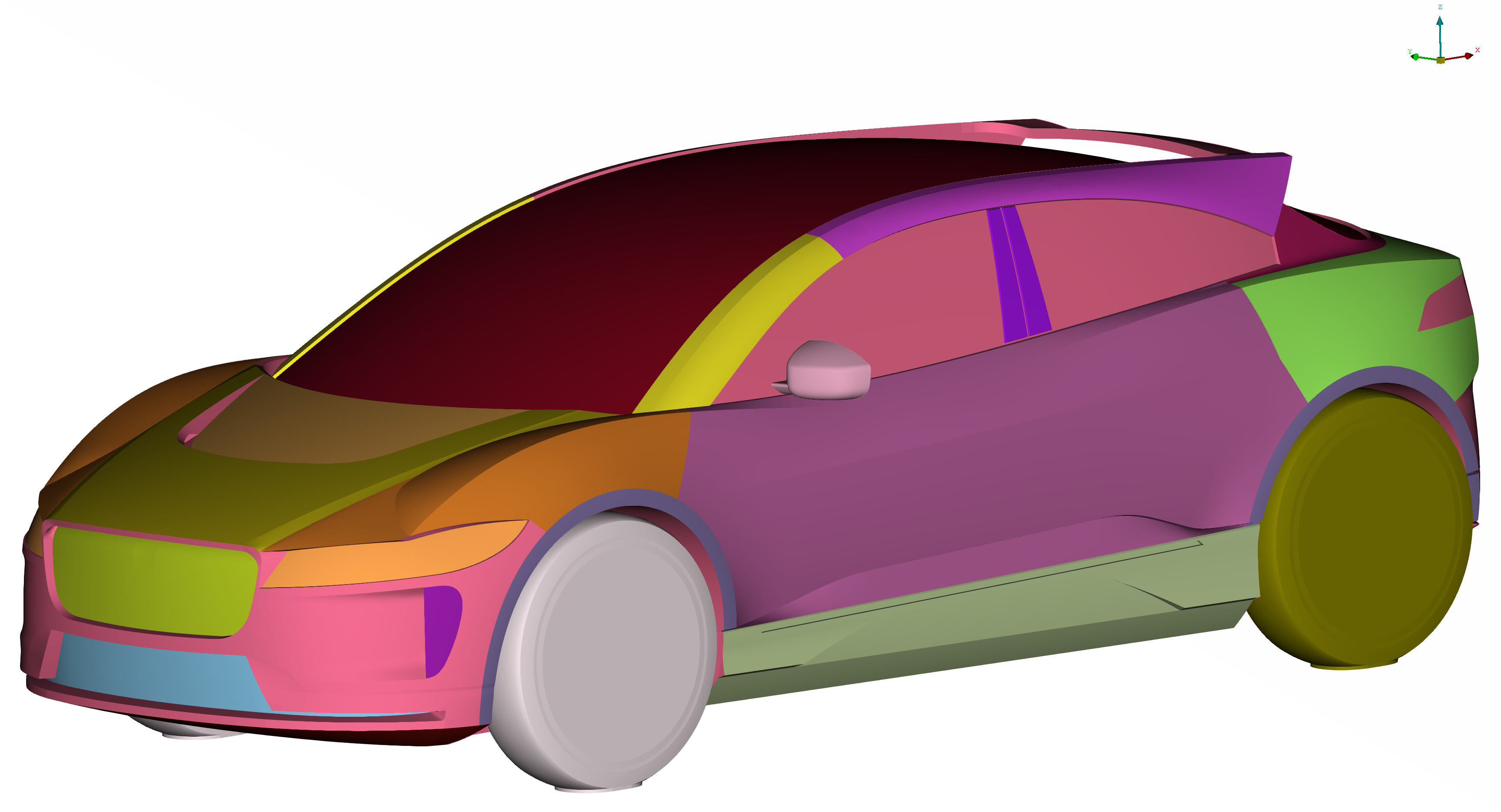}
        \caption*{C. Self-driving cars}
    \end{minipage}
    \caption{PCV Models Sample Showcase}
    \label{fig:PCV}
\end{figure*}

\subsection{Dynamic learnable integration} 
While edge convolution captures local spatial features of point clouds, the correlation attention module proposed in this study provides global semantic information. To improve the robustness of their fusion, learnable parameters \(\alpha\), initialized with specific values, are employed to adaptively weight the integration of local and global features. The fusion process can be formulated as follows:
`
\begin{align}
G^v = AconC(\gamma) \\
G^o = \alpha \cdot A \cdot G^v + \gamma
\end{align}

\noindent where \(G^o\) represents the output features extracted by the regression prediction network. The AconC \cite{ma2021activate} activation function introduces a switching factor to learn the parameter transition between non-linear (activation) and linear (non-activation) states. This allows the model to more flexibly select the appropriate mapping approach based on varying point cloud features.

\subsection{Training and Evaluation of Models} 

The training of the DAT model adopts Mean Squared Error (MSE) as the primary loss function, combined with the Adam optimization algorithm \cite{kingma2014adam} for parameter optimization. During the training process, strategies such as learning rate adjustment and L2 Regularization (or Weight Decay) are employed to prevent model overfitting. 

The model's performance is evaluated using metrics including \(R^2\) score (Coefficient of Determination), Mean Absolute Percentage Error (MAE), and Mean Squared Error (MSE), which comprehensively reflect the model's predictive capability and robustness. To accurately assess the model's performance, the following three evaluation metrics are used: MSE, MAE, and \(R^2\), as shown in Equations (11) to (13).
\begin{align}
\text{MSE} &= \frac{1}{n} \sum_{i=1}^{n} (y_i - y_i')^2 \\
\text{MAE} &= \frac{1}{n} \sum_{i=1}^{n} \left| y_i - y_i'\right| \\
R^2 &= 1 - \frac{\sum_{i=1}^{n} (y_i - y_i')^2}{\sum_{i=1}^{n} (y_i - \bar{y})^2}
\end{align}
where:
\(y_i\) represents the ground truth value, and \(y_i'\) is the predicted value.

The closer MSE and MAE are to 0, and the closer \(R^2\) is to 1, the higher the prediction accuracy of the model.

\section{Validation and Analysis}
\subsection{Additions to the DrivAerNet++ validation set}

With China's rapid economic development, consumer demand for large-space models (e.g., SUVs, crossover ORVs, and multi-purpose vehicles MPVs) continues to grow \cite{du2018insights, helveston2015will}. However, the original DrivAerNet++ dataset provides insufficient coverage of these vehicle categories and does not fully reflect trends in the Chinese automotive market. To improve model adaptability and generalization, this study supplements the training set with additional SUV and MPV models, addressing the lack of diversity in the validation data.

Since DrivAerNet++ is constructed based on the original DrivAer model, this paper only adds some new models in the PCV (common passenger vehicle) part for validation, and the rest of the main models are obtained by sampling the existing dataset. The selected part of SUV, ORV, MPV and PCV models are shown in Figures ~\ref{fig:SUV} and ~\ref{fig:PCV}.

According to results from AutoCFD1, methods such as RANS, WMLES (Wall-Modelled Large Eddy Simulation), and hybrid RANS-LES (HRLM) exhibit significant deviation from wind tunnel experiments~\cite{ashton2022overview}. Therefore, this study uses wind tunnel experimental data—rather than CFD simulations—for constructing the validation set, with some models provided by collaborating automotive companies.

\begin{figure}[h]
    \centering
    \includegraphics[width=0.43\textwidth]{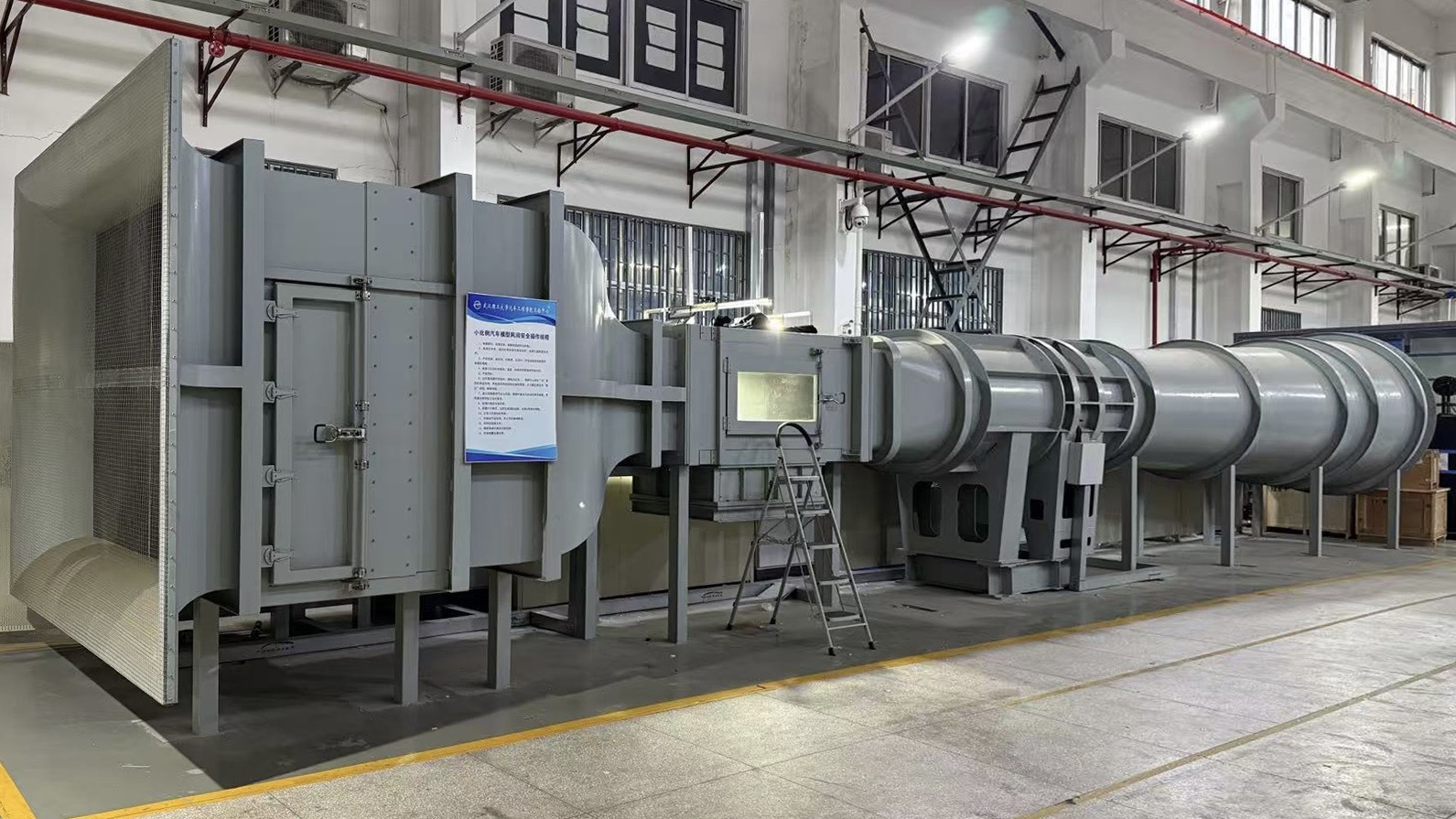}
    \caption{Wind Tunnel}
    \label{fig:wind_tunnel}
\end{figure}

\begin{figure}[h]
    \centering
    \includegraphics[width=0.43\textwidth]{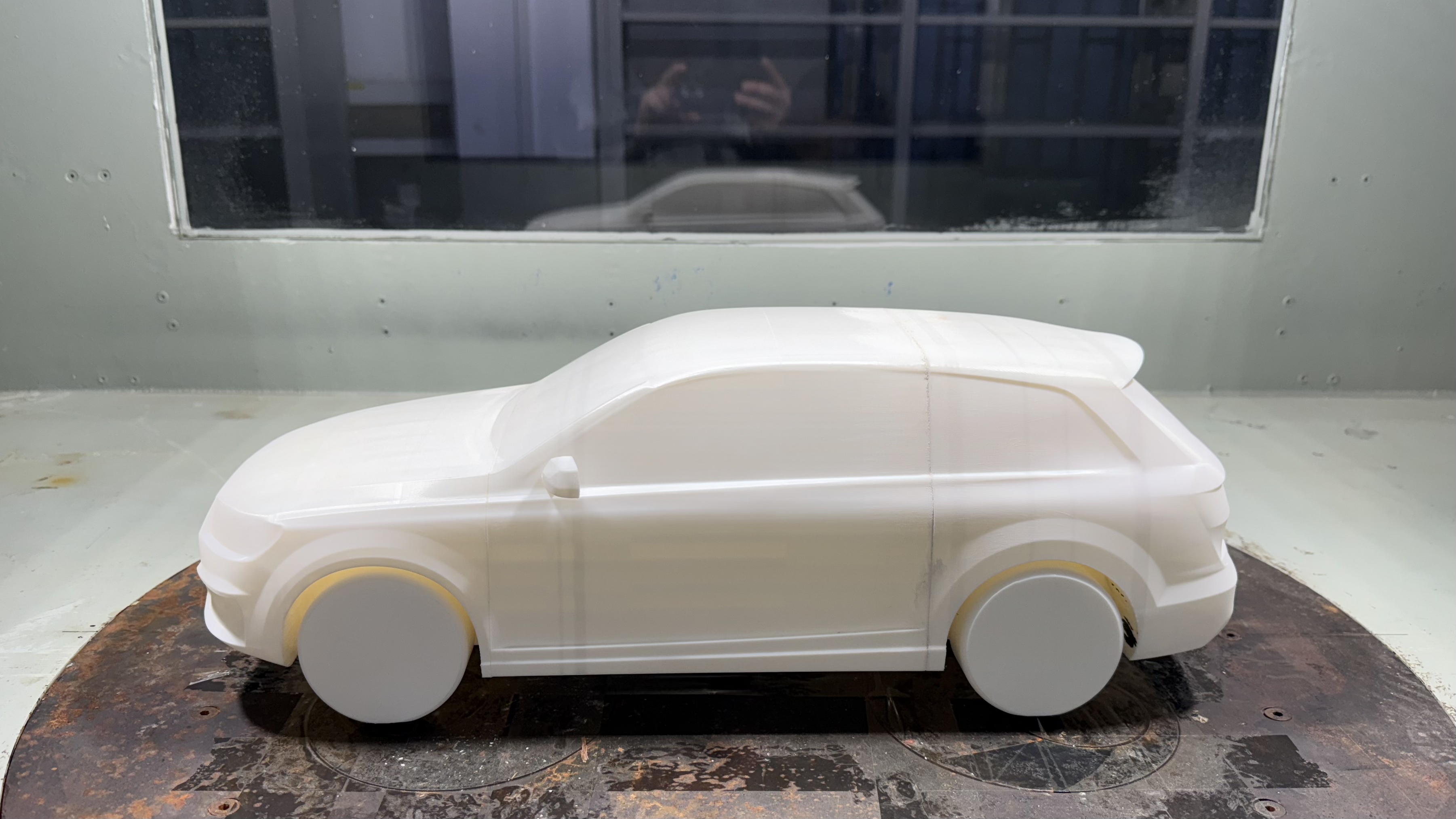}
    \caption{Wind Tunnel Testing}
    \label{fig:wind_tunnel_test}
\end{figure}

\begin{figure*}[t]
    \centering
    \begin{subfigure}{0.49\textwidth}
        \centering
        \includegraphics[width=\textwidth]{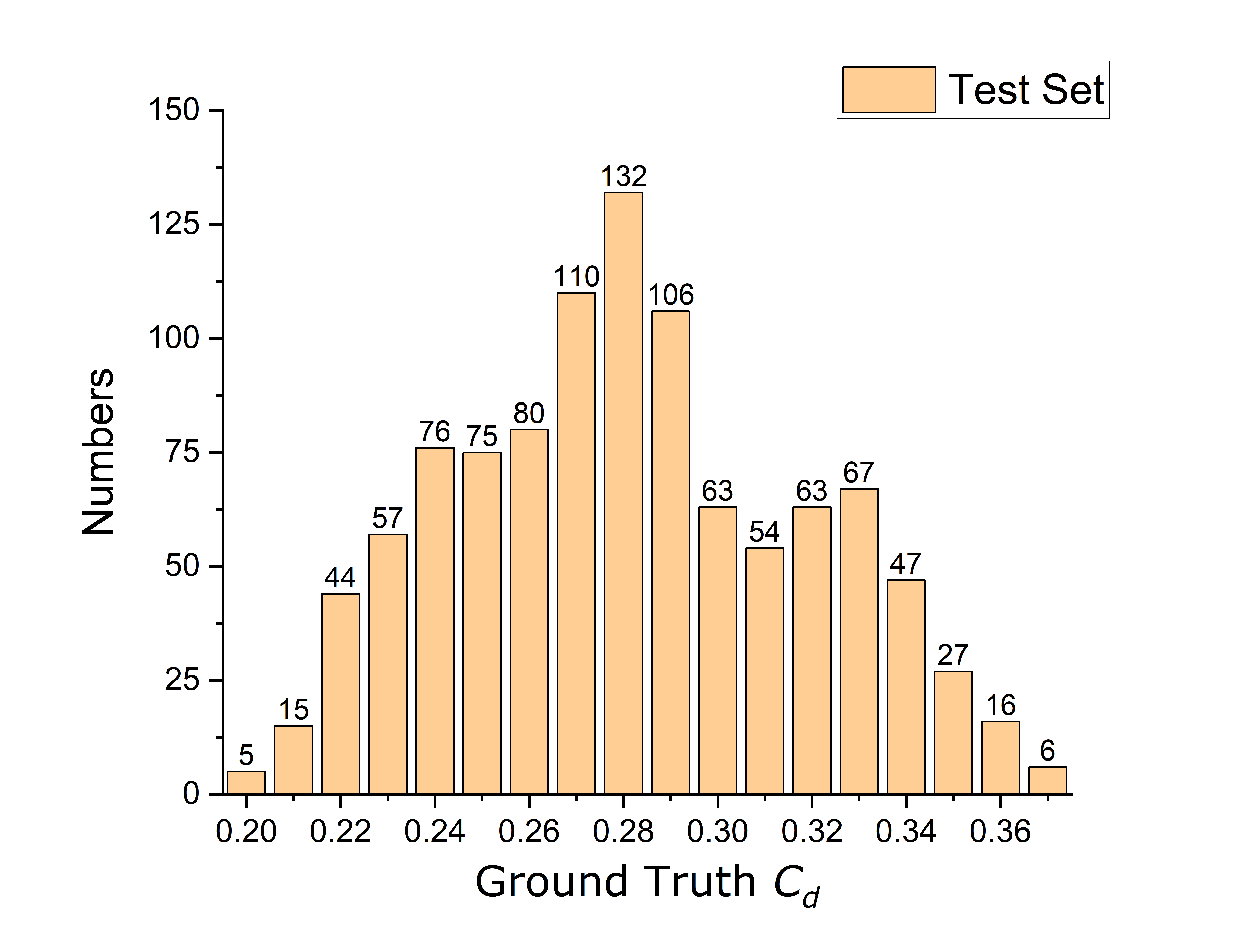}
        \caption{Distribution of ground truth drag coefficients in the DrivAerNet++ test set.}
    \end{subfigure}
    \hfill
    \begin{subfigure}{0.49\textwidth}
        \centering
        \includegraphics[width=\textwidth]{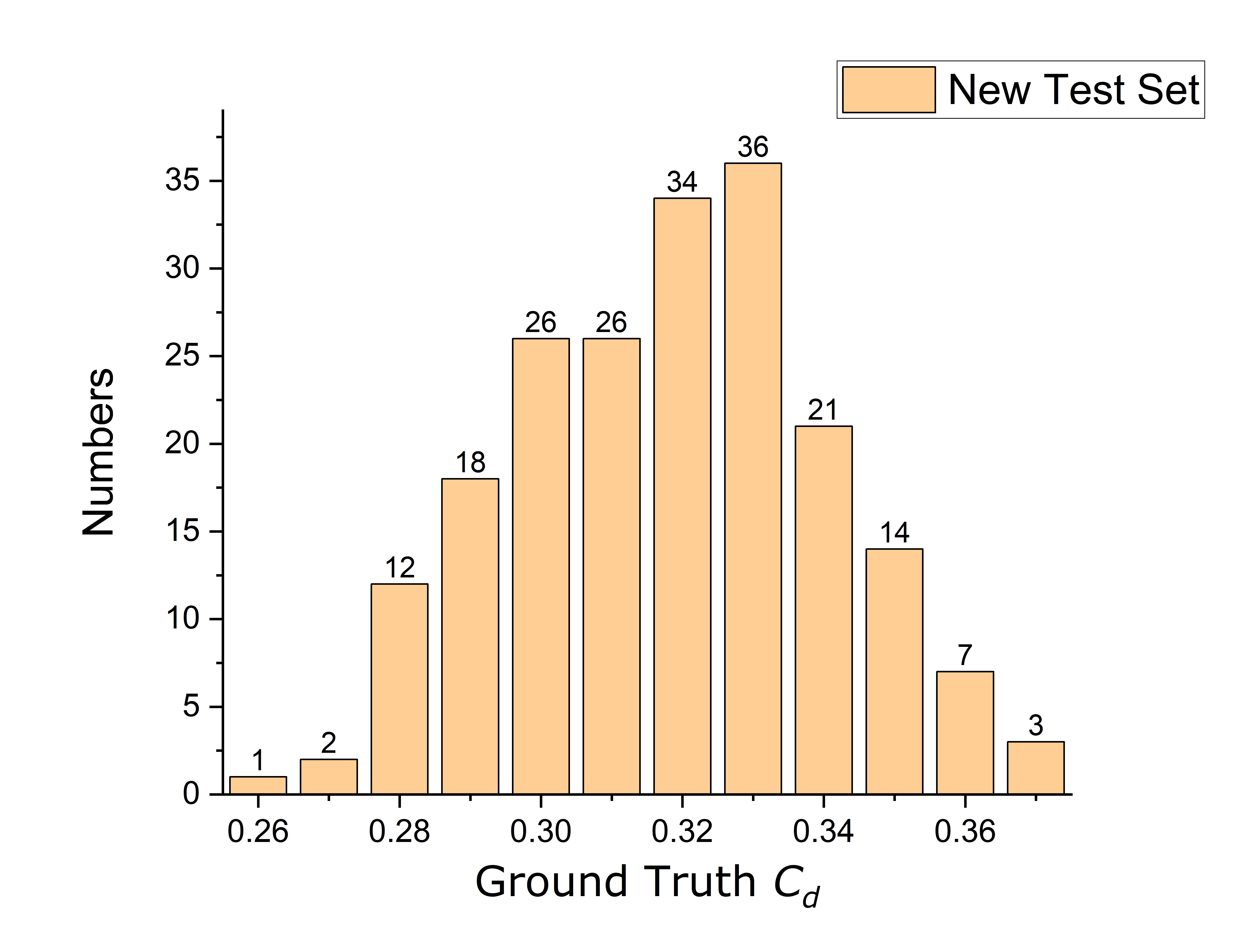}
        \caption{The drag coefficients in new validation set}
    \end{subfigure}
    \caption{Distribution of ground truth drag coefficients of two test Set}
    \label{fig:validation_sets}
\end{figure*}

This wind tunnel-centered validation dataset improves realism and model applicability. It is built by combining existing and newly acquired experimental data under cost constraints. Experiments are conducted at the Hubei Provincial Key Laboratory of Automotive Components Technology, Wuhan University of Technology. Equipment configuration and testing conditions are shown in Fig.~\ref{fig:wind_tunnel}. 

During testing, vehicles were fixed in position to ensure consistency, tires were kept stationary to avoid interference, and body seams were polished to minimize surface resistance. These controls ensured the accuracy and repeatability of results, as illustrated in Fig.~\ref{fig:wind_tunnel_test}.

\begin{figure}[h]
    \centering
    \includegraphics[width=0.49\textwidth]{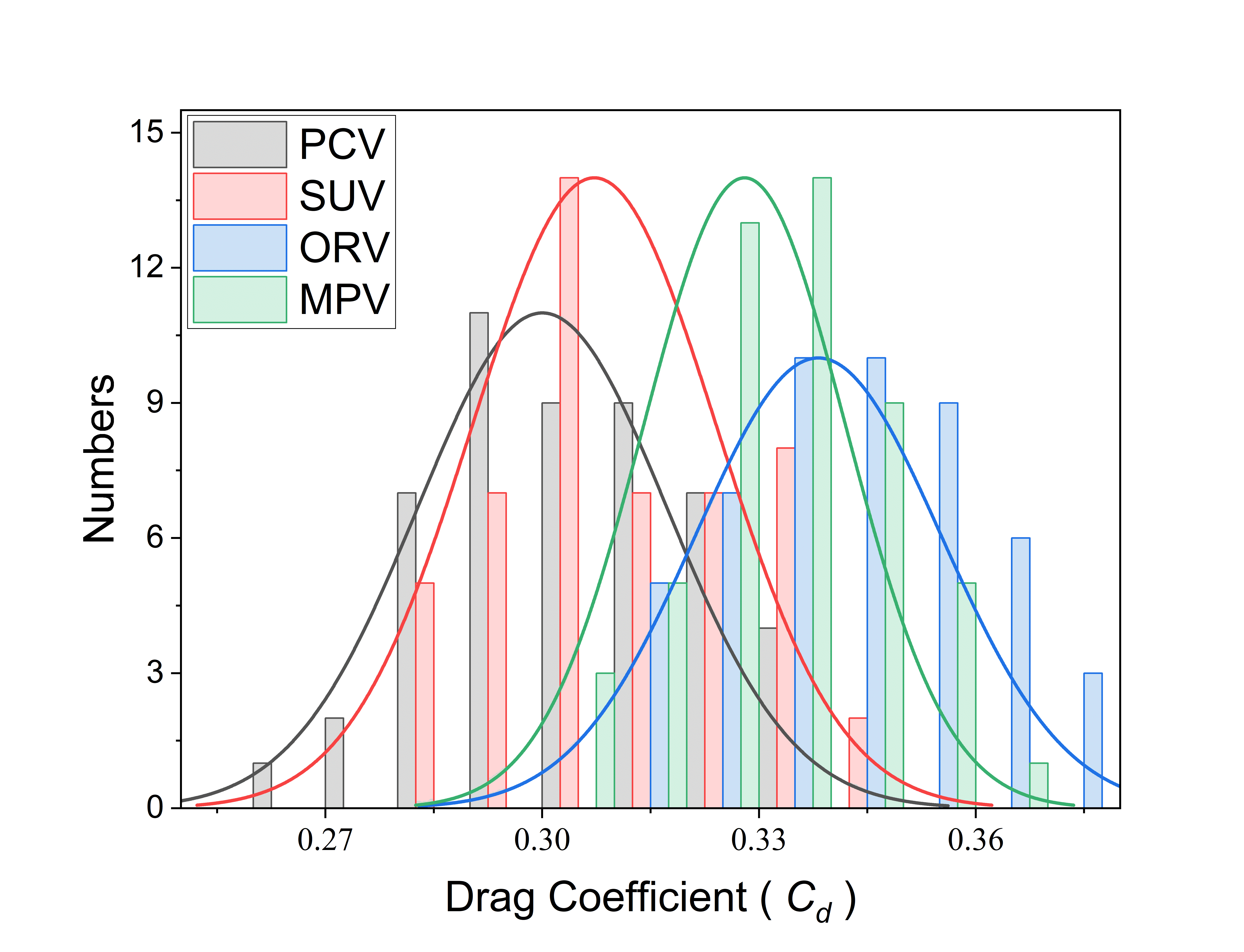}
    \caption{Normal distribution of New Validation Set}
    \label{fig:normal}
\end{figure}

As shown in Fig.~\ref{fig:validation_sets}(a), the drag coefficients ($C_d$) in the DrivAerNet++ validation set are uniformly distributed between 0.2 and 0.4. The extended validation set (Fig.~\ref{fig:validation_sets}(b) and Fig.~\ref{fig:normal}) shows a concentration of samples in the 0.30--0.34 range, which aligns with the aerodynamic characteristics of mainstream SUVs and MPVs~\cite{ekman2020assessment, wang2019effect, windsor2014real}. This extension enhances generalization to different vehicle types and provides targeted support for applying the DAT architecture in the Chinese market.

\section{Results and discussion}
\subsection{Implementation Details}
All experiments herein were performed on a computer equipped with two AMD EPYC 9K84 CPUs and four NVIDIA H20 GPUs. The operating system used was Ubuntu 20.04, and the experiments were implemented using Python 3.8. The deep learning framework employed was PyTorch 2.0.0, optimised with CUDA 11.8 and cuDNN 8.7.0 to ensure efficient computation. All computations, including model training and inference, were accelerated using the GPU and CUDA to achieve optimal performance and reduced processing time.

The hyperparameter settings for the model included an initial learning rate of $10^{-4}$, with the optimizer used being adaptive moment estimation (Adam)\cite{kingma2014adam}. The number of nearest neighbors for graph construction was 50, and the number of points in the point cloud was 5000. The overall network was trained for 100 epochs with a batch size of 32 under 15 hours.

\subsection{Results on DrivAerNet++ Test Set}

\begin{figure*}[t]
    \centering
    \begin{subfigure}{0.49\textwidth}
        \centering
        \includegraphics[width=\textwidth]{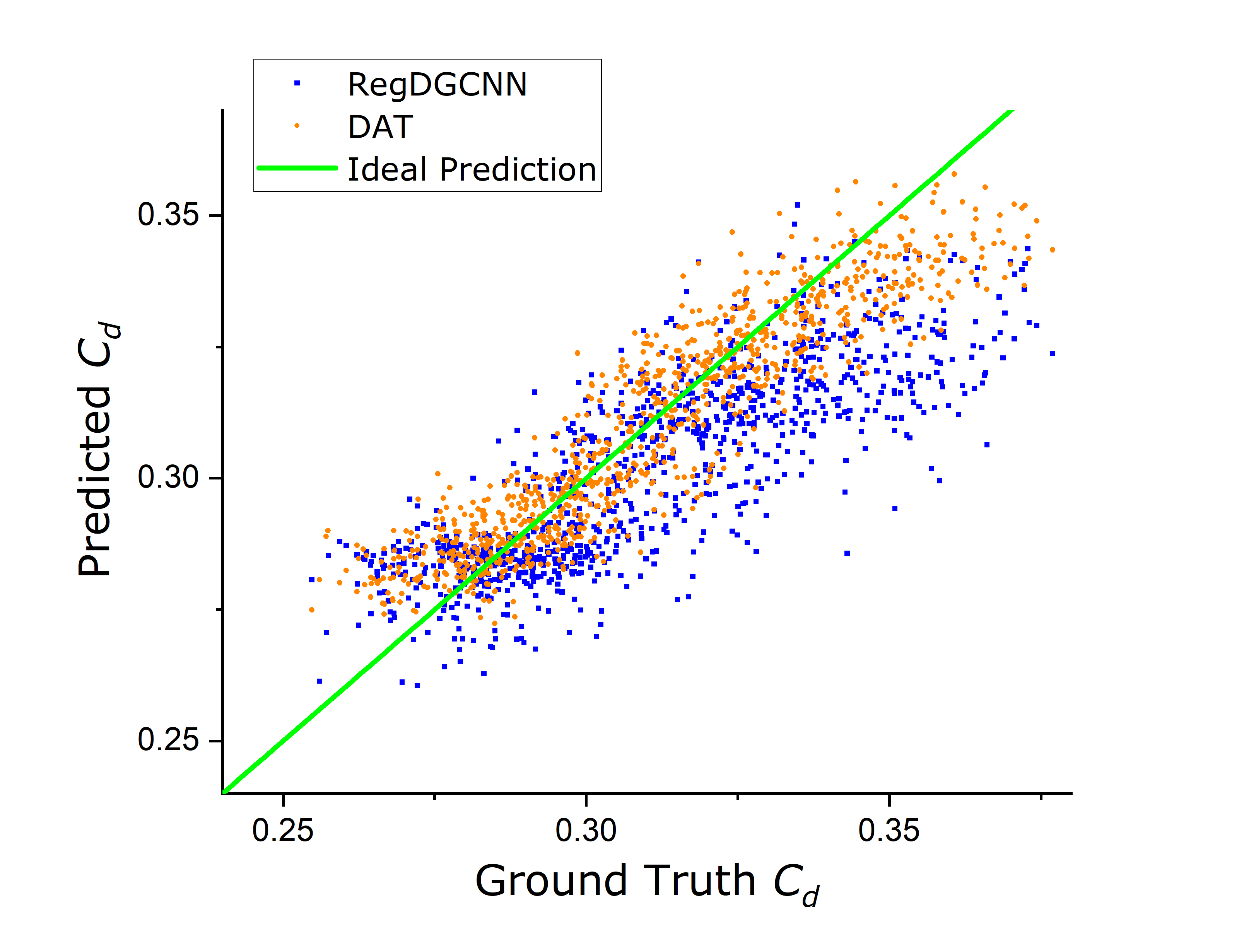}
        \caption{Correlation of Predicted vs. Ground Truth \(C_d\)}
    \end{subfigure}
    \hfill
    \begin{subfigure}{0.49\textwidth}
        \centering
        \includegraphics[width=\textwidth]{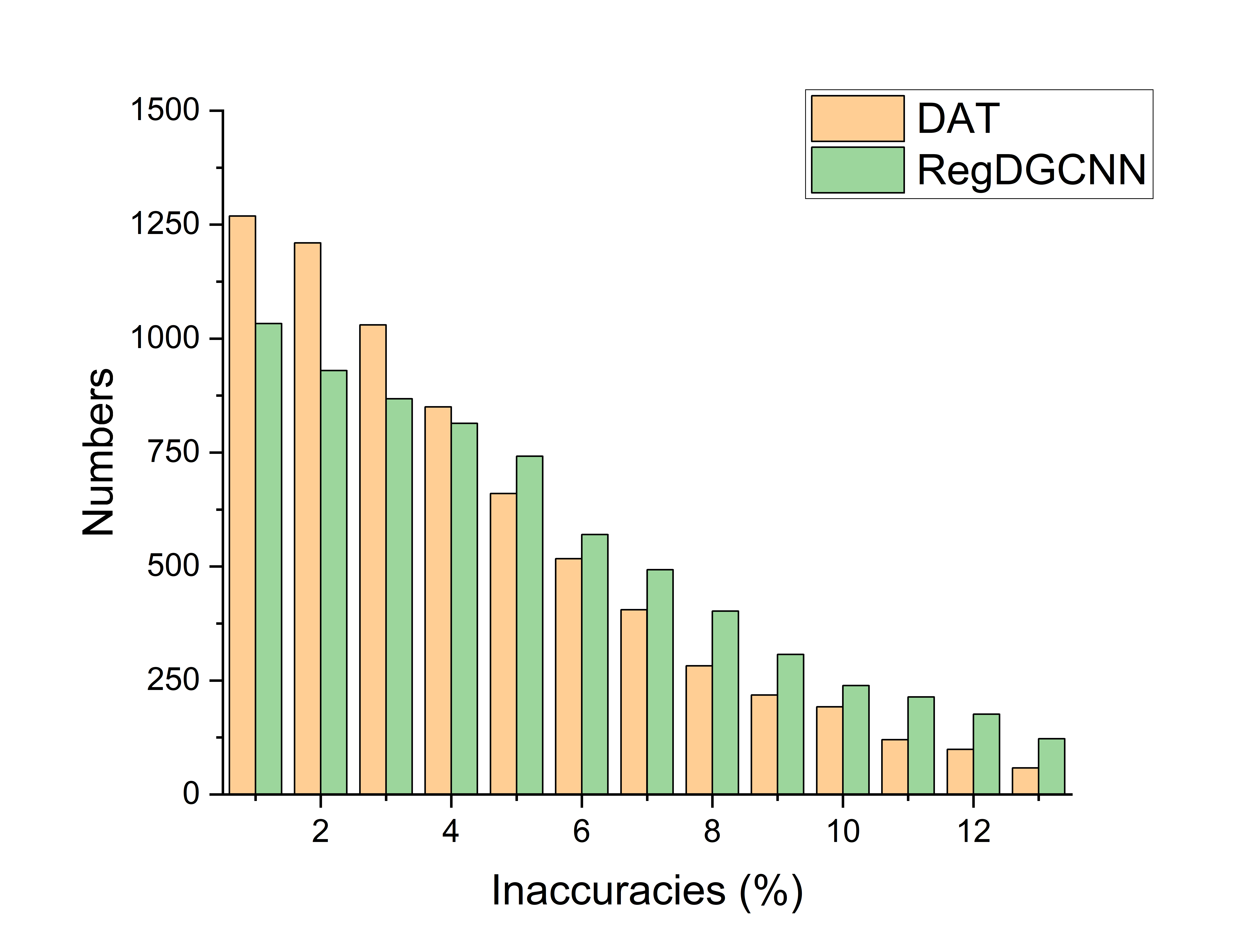}
        \caption{DrivAerNet++ Test Set's Error Distribution}
    \end{subfigure}
    \caption{Error Analysis on the DrivAerNet++ Test Set}
    \label{fig:ed}
\end{figure*}

The test results of PointNet~\cite{qi2017pointnet}, GCNN~\cite{kipf2016semi}, RegDGCNN~\cite{elrefaie2024drivaernet++}, Transolver~\cite{wu2024transolver}, Transolver++~\cite{luo2025transolver++} and DAT under the validation set of DrivAerNet++ are shown in Tab.~\ref{tab:results} and~\ref{tab:results2}. It can be seen that although the MSE and MAE values of the first three baseline methods are relatively close, DAT achieves a significantly higher $R^2$ score (0.872), indicating a better overall regression performance. As shown in Tab.~\ref{tab:results}, although Transolver and Transolver++ exhibit reasonable performance in terms of MSE and MAE, their $R^2$ values (0.5712 and 0.6719, respectively) remain substantially lower than that of DAT, revealing inferior consistency between predicted and true drag values.

\begin{table}[h]
\caption{\label{tab:results}Comparison of regression performance on the DrivAerNet++ validation set}
\begin{ruledtabular}
\begin{tabular}{lcccc}
\textbf{Model} & \textbf{MSE} & \textbf{MAE} & \textbf{Max AE} & \textbf{$R^2$} \\
\hline
PointNet\cite{qi2017pointnet}           & 0.000149  & 0.00960  & 0.01245  & 0.644 \\
GCNN\cite{kipf2016semi}                 & 0.000171  & 0.01043  & 0.01503  & 0.596 \\
RegDGCNN\cite{elrefaie2024drivaernet++}   & 0.000142  & 0.00931  & 0.01279  & 0.643 \\
Transolver\cite{wu2024transolver}       & 0.000603  & 0.02031  & 0.06561  & 0.571 \\
Transolver++\cite{luo2025transolver++}  & 0.000461  & 0.01769  & 0.05795  & 0.672 \\
DAT                                     & 0.000118  & 0.00911  & 0.01126  & 0.872 \\
\end{tabular}
\end{ruledtabular}
\end{table}

\begin{table}[h]
\caption{\label{tab:results2}Comparison of Model Efficiency and Complexity}
\begin{ruledtabular}
\begin{tabular}{lccc}
\textbf{Model} & \textbf{Training Time} & \textbf{Inference Time}  & \textbf{Flops}\\
\hline
PointNet\cite{qi2017pointnet}           & 4.7hrs   & 0.84s  & 474.98G \\
GCNN\cite{kipf2016semi}                 & 49hrs    & 50.8s  & 260.24G \\
RegDGCNN\cite{elrefaie2024drivaernet++}   & 12.6hrs  & 0.85s  & 474.29G \\
Transolver\cite{wu2024transolver}       & 45.7hrs  & 0.66s  & 214.31G \\
Transolver++\cite{luo2025transolver++}  & 46.1hrs  & 0.63s  & 214.41G \\
DAT                                     & 14.3hrs  & 0.73s  & 488.11G \\
\end{tabular}
\end{ruledtabular}
\end{table}

From the computational efficiency perspective (Tab.~\ref{tab:results2}), it is evident that models like PointNet and RegDGCNN have comparable inference times (0.84s and 0.85s, respectively), while GCNN shows significantly slower inference (50.8s), due to its graph processing overhead. Although DAT has a slightly higher FLOPs value (488.11G), its inference time (0.73s) remains competitive, thanks to the parallel-friendly structure of the attention mechanism. Transolver++ has the fastest inference (0.63s) with relatively lower FLOPs (214.41G), but it comes at the cost of longer training time (46.1 hours). In contrast, DAT requires only 14.3 hours for training, demonstrating a favorable balance between performance and efficiency.

The superior accuracy of DAT is primarily attributed to the Correlation-Driven Attention module, which effectively captures long-range dependencies and geometric correlations in unordered 3D point clouds. This attention mechanism allows the model to better generalize across different body shapes and flow patterns. Furthermore, the correlation-based feature weighting and the estimator module improve the representation of spatially critical regions, leading to lower overall prediction error.

After evaluating the validation set, the proposed DAT architecture exhibits superior adaptability across a wide range of vehicle types. The incorporation of the self-attention mechanism~\cite{vaswani2017attention} enables the model to capture long-range dependencies and global geometric semantics, thereby enhancing predictive accuracy, particularly in scenarios involving high geometric variability. As illustrated in Fig.~\ref{fig:ed}, DAT predictions are highly concentrated within the 0–4\% error interval, while RegDGCNN produces a more uniform error distribution. This concentration indicates not only lower average prediction error but also improved consistency and robustness across samples. The full prediction results across all subsets of the DrivAerNet++ test set are visualized in the Supplementary Material (Figs.~\ref{fig:Material}), further illustrating the consistency and coverage of DAT across diverse configurations.

\begin{figure*}[t]
    \centering
    \begin{subfigure}{0.48\textwidth}
        \centering
        \includegraphics[width=\textwidth]{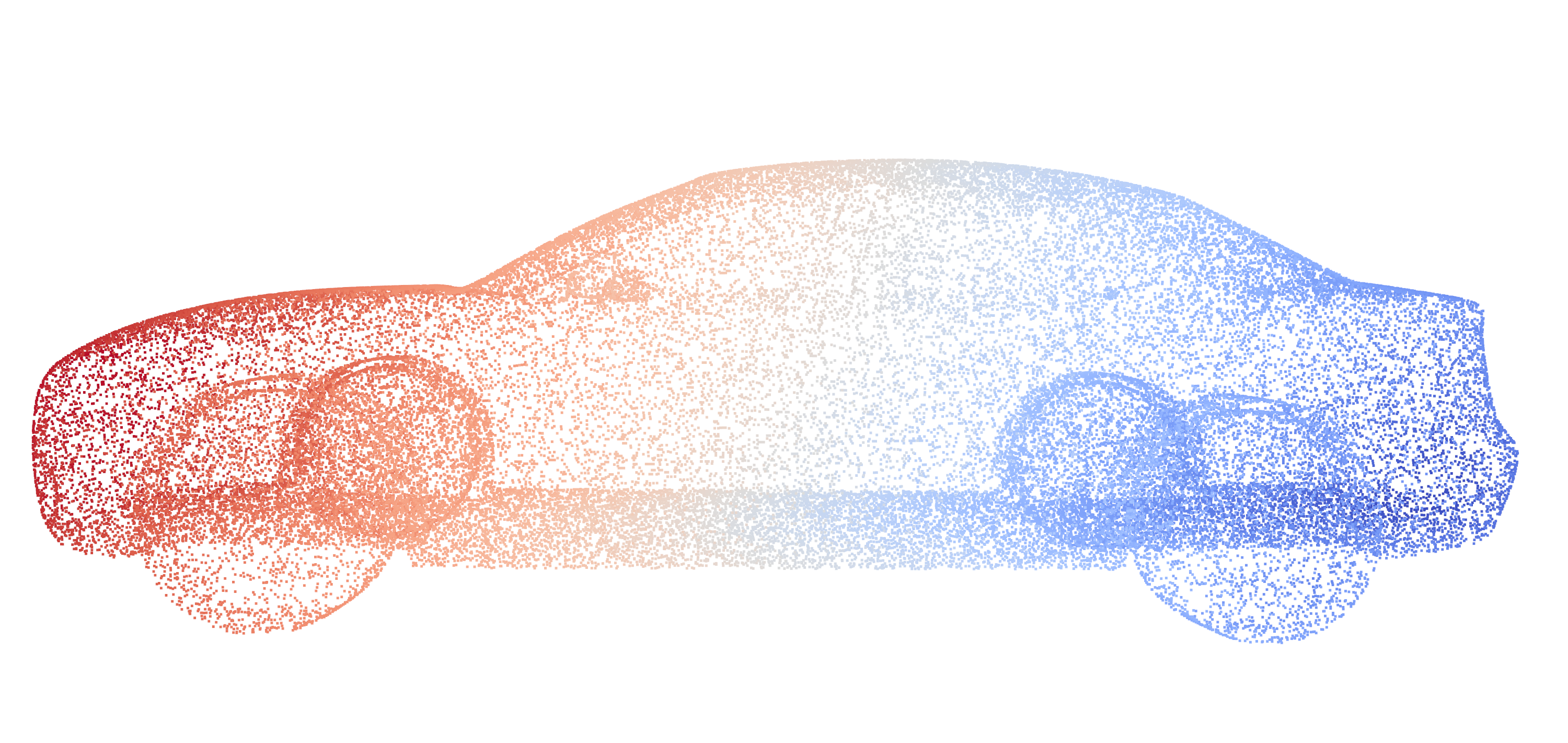}
        \caption{PCV - XY Plane View}
    \end{subfigure}
    \hfill
    \begin{subfigure}{0.48\textwidth}
        \centering
        \includegraphics[width=\textwidth]{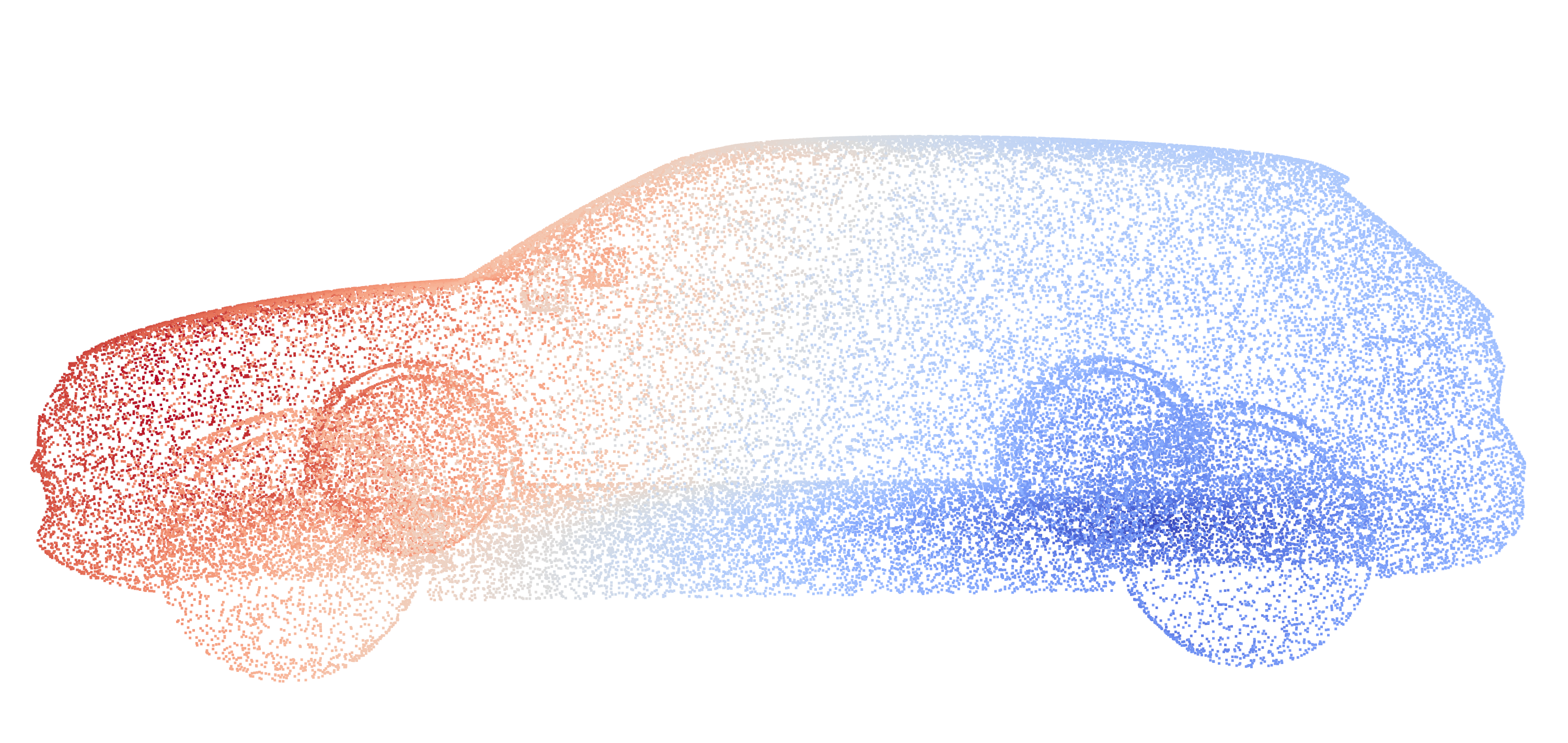}
        \caption{SUV - XY Plane View}
    \end{subfigure}
    \vskip 1em
    \begin{subfigure}{0.4\textwidth}
        \centering
        \includegraphics[width=\textwidth]{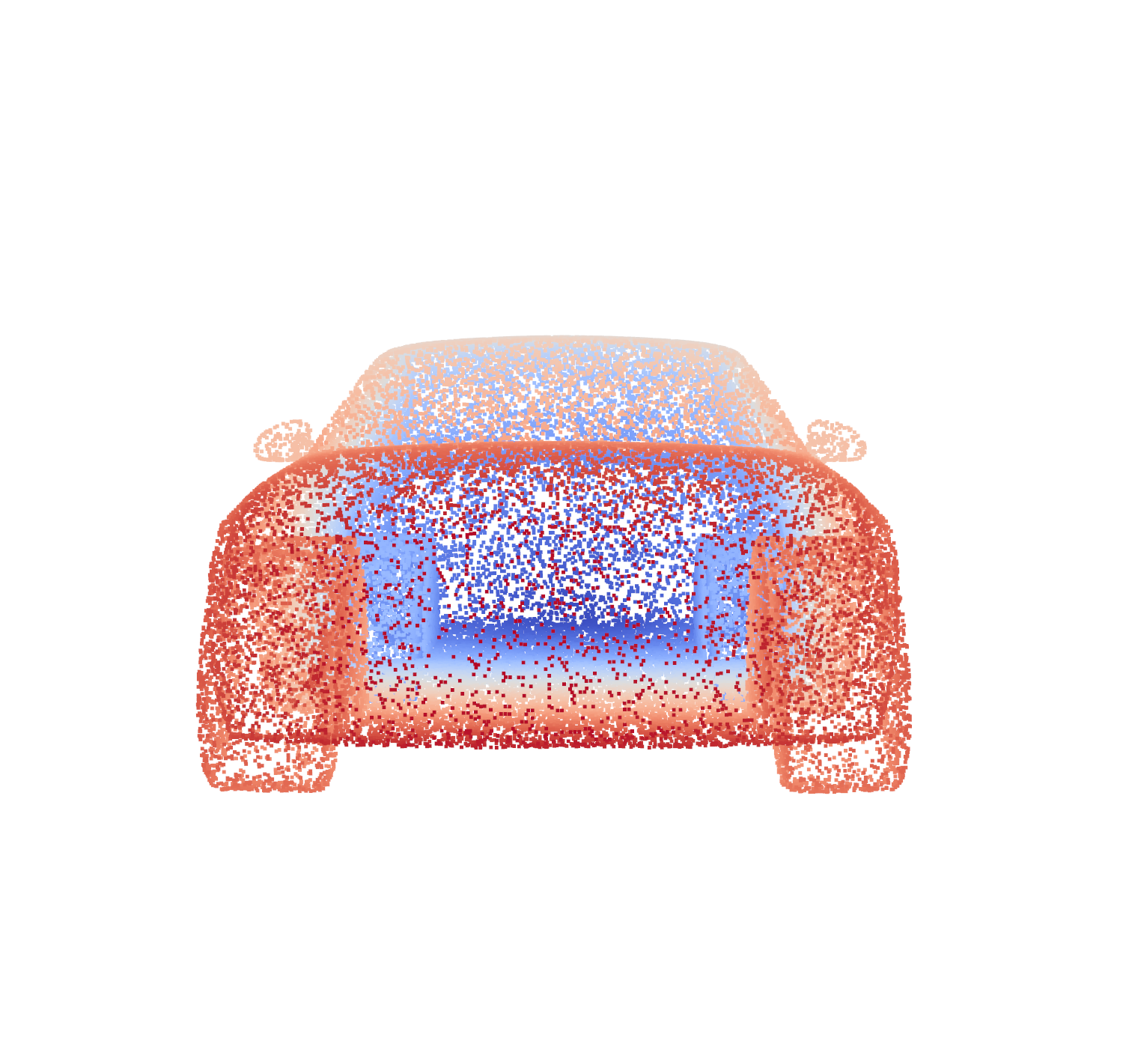}
        \caption{PCV - YZ Plane View}
    \end{subfigure}
    \hfill
    \begin{subfigure}{0.4\textwidth}
        \centering
        \includegraphics[width=\textwidth]{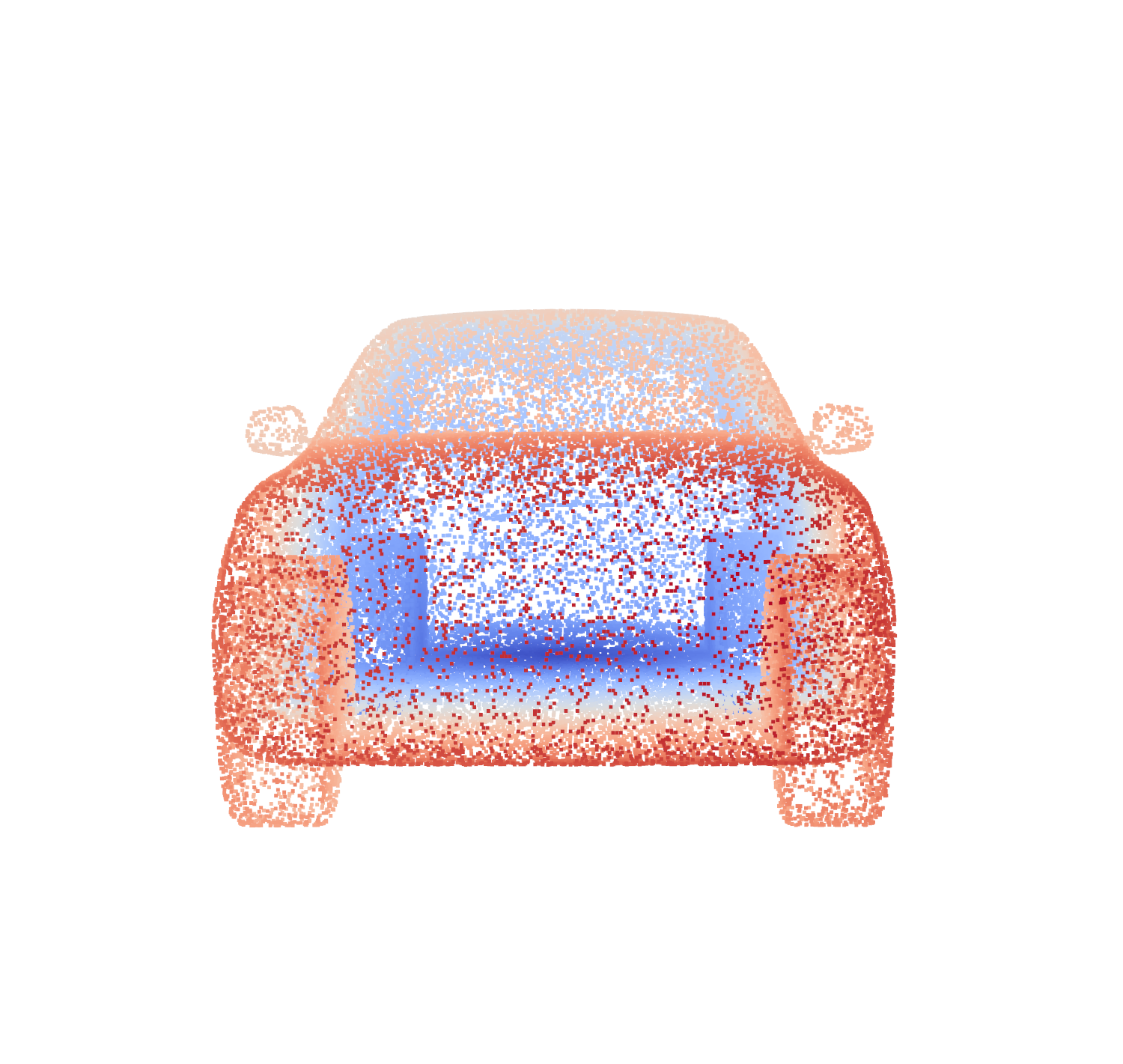}
        \caption{SUV - YZ Plane View}
    \end{subfigure}
    \caption{Attention maps on the XY and YZ planes (red = high attention, blue = low)}
    \label{fig:attention}
\end{figure*}

\begin{figure*}[t]
    \centering
    \begin{subfigure}{0.49\textwidth}
        \centering
        \includegraphics[width=\textwidth]{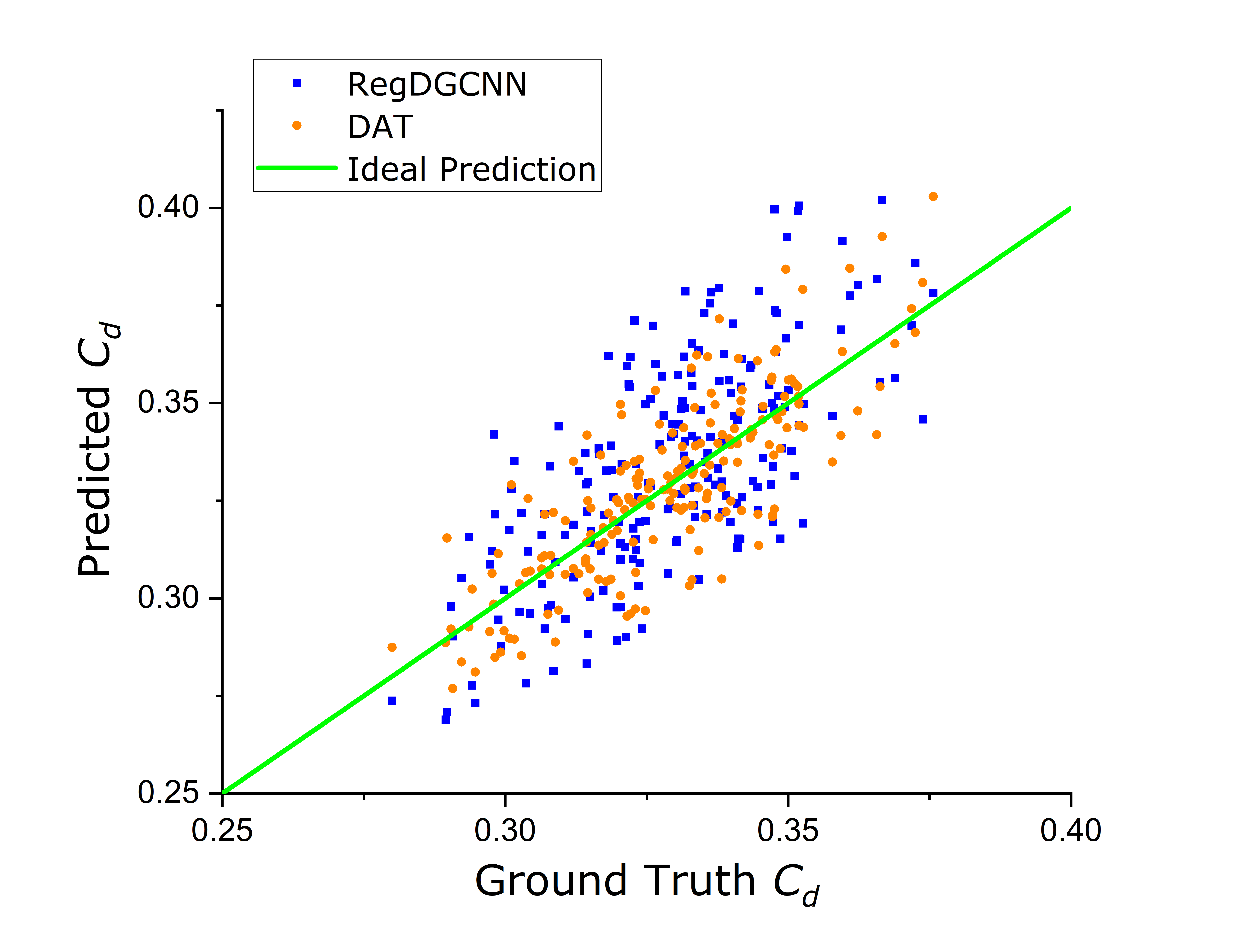}
        \caption{Correlation of Predicted vs. Ground Truth \(C_d\)}
    \end{subfigure}
    \hfill
    \begin{subfigure}{0.49\textwidth}
        \centering
        \includegraphics[width=\textwidth]{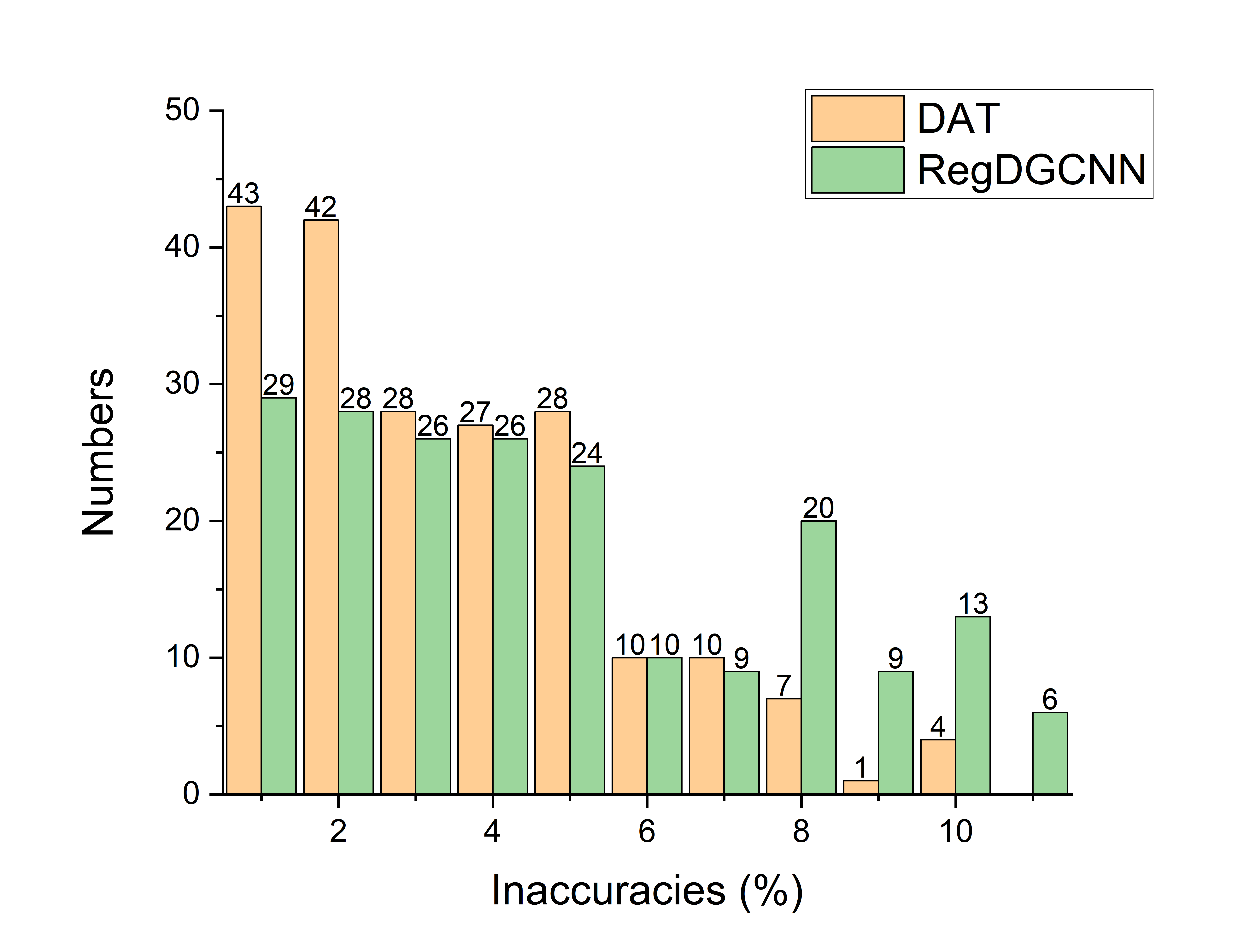}
        \caption{Error Distribution on the New Test Set}
    \end{subfigure}
    \caption{Error Analysis on the New Test Set}
    \label{fig:new ed}
\end{figure*}

\subsection{Ablation Study}
To evaluate the effectiveness of the proposed DrivAer Transformer (DAT) and its core components, a comprehensive ablation study was conducted. These experiments were designed to quantify the contribution of each module to the model’s overall performance. As shown in Table~\ref{tab:ablation}, RegDGCNN was adopted as the baseline model, and the impact of various components and design strategies on DAT was systematically analyzed.

\begin{table}[htbp]
\caption{Ablation Study Results of DAT Components}
\centering
\begin{ruledtabular}
\begin{tabular}{lccc}
\textbf{Model} & \textbf{$R^2$} & \textbf{FLOPs} & \textbf{Parameters} \\
\hline
RegDGCNN (baseline)              & 0.643 & 474.29 G & 3.17 M \\
Baseline + CDA                  & 0.786 & 487.73 G & 5.41 M \\
Baseline + CDA + CDE (DAT)      & \textbf{0.872} & 488.11 G & 5.41 M \\
DAT without learnable variables & 0.859 & 488.11 G & 5.41 M \\
DAT without residuals           & 0.826 & 488.11 G & 5.41 M \\
\end{tabular}
\label{tab:ablation}
\end{ruledtabular}
\end{table}

The experiments were performed on the DrivAerNet++ dataset, using the coefficient of determination ($R^2$) as the primary performance metric. Additionally, the number of floating-point operations (FLOPs) and model parameters were reported to reflect computational cost and memory complexity.

Introducing the Correlation-Driven Attention (CDA) module increased the FLOPs by 13.44G and parameter count by 2.24M, while significantly improving the $R^2$ score by 0.143. This confirms the module’s capability in capturing global geometric correlations in point cloud representations. Further improvement was observed by incorporating the Correlation-Driven Estimator (CDE), which boosted the $R^2$ score by an additional 0.086. This enhancement is attributed to the integration of local similarity features, obtained from CDA, into global structural representations, which aids in learning category-specific vehicle geometry patterns.

Moreover, the removal of learnable parameters and residual connections led to a decrease in $R^2$ by 0.013 and 0.046, respectively. These findings highlight the importance of attention-branch control and residual pathways in stabilizing the training process and accelerating model convergence.

To further validate the proposed architecture, we visualize the learned attention distributions on the vehicle surface (see Figs.~\ref{fig:attention}). The attention values are derived from the similarity matrix in the CDA module, where red indicates high importance and blue indicates low importance.

The results show that the DAT model consistently attends to key aerodynamic regions, especially around the front bumper and windshield slope, which are known to strongly influence drag. This confirms that the attention mechanism effectively captures critical geometric features relevant to aerodynamic performance.

\subsection{Results on New Test Set}

\begin{figure*}[t]
    \centering
    \begin{subfigure}{0.43\textwidth}
        \centering
        \includegraphics[width=\textwidth]{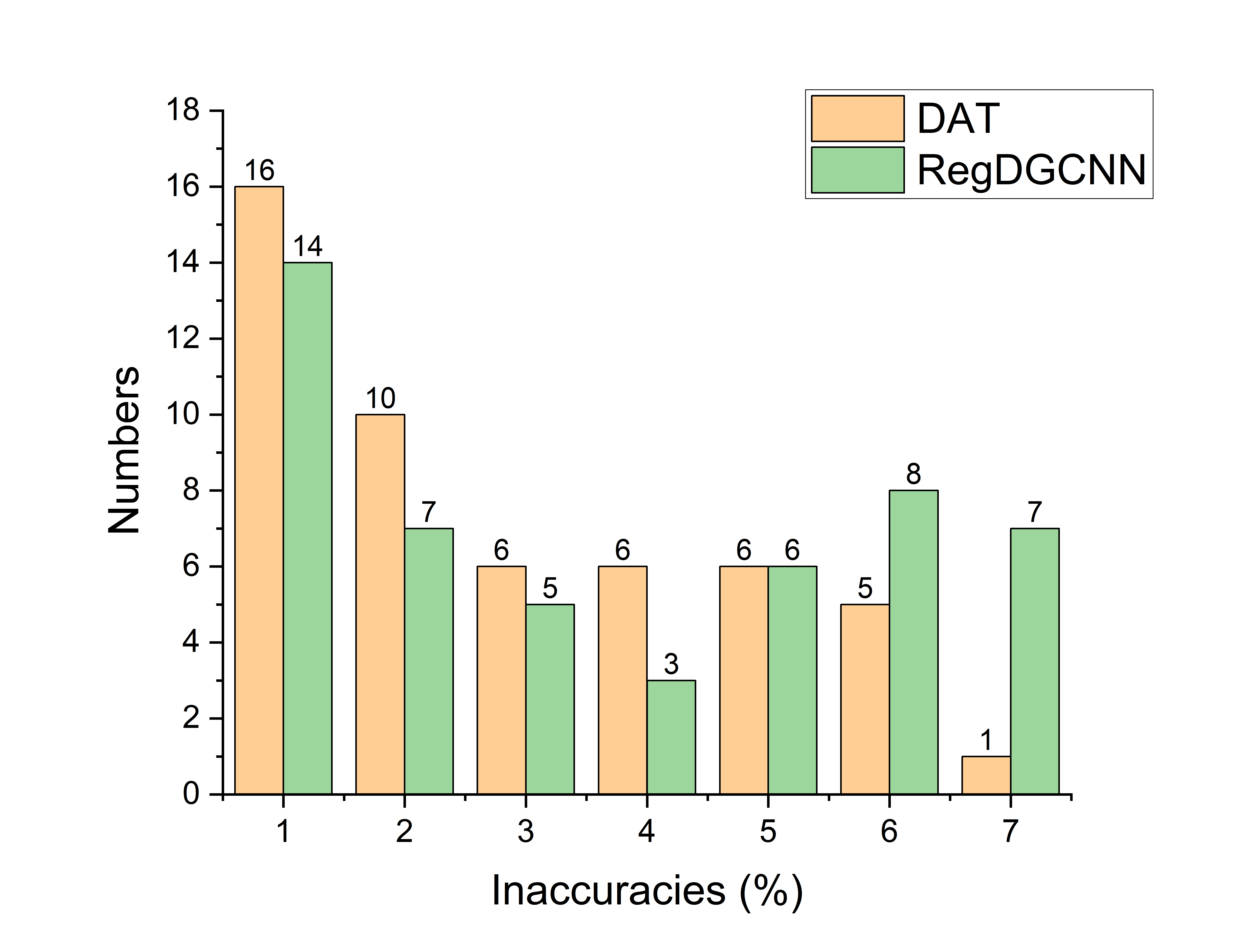}
        \caption{PCV Error Distribution}
    \end{subfigure}
    \hfill
    \begin{subfigure}{0.43\textwidth}
        \centering
        \includegraphics[width=\textwidth]{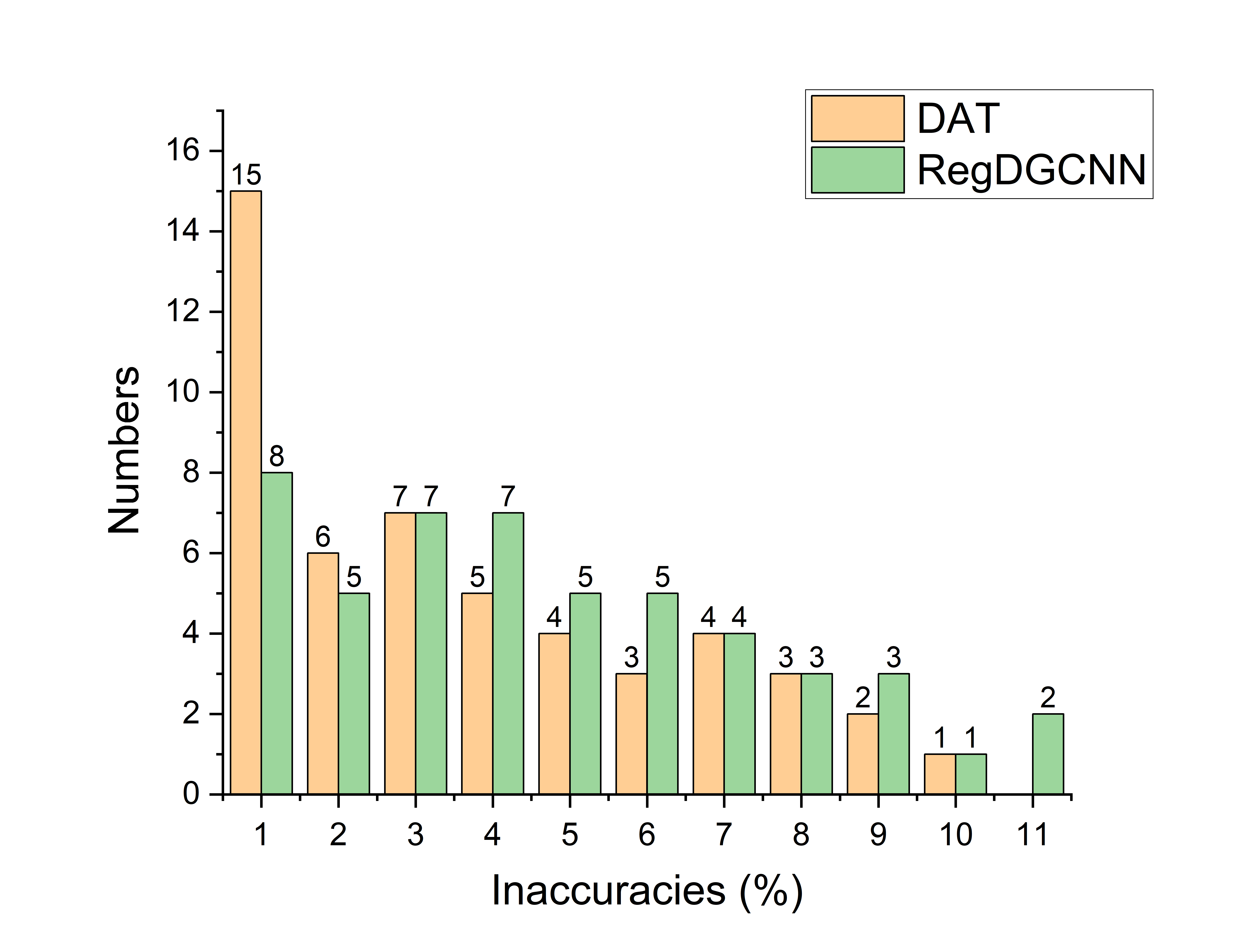}
        \caption{SUV Error Distribution}
    \end{subfigure}
    \vskip 1em
    \begin{subfigure}{0.43\textwidth}
        \centering
        \includegraphics[width=\textwidth]{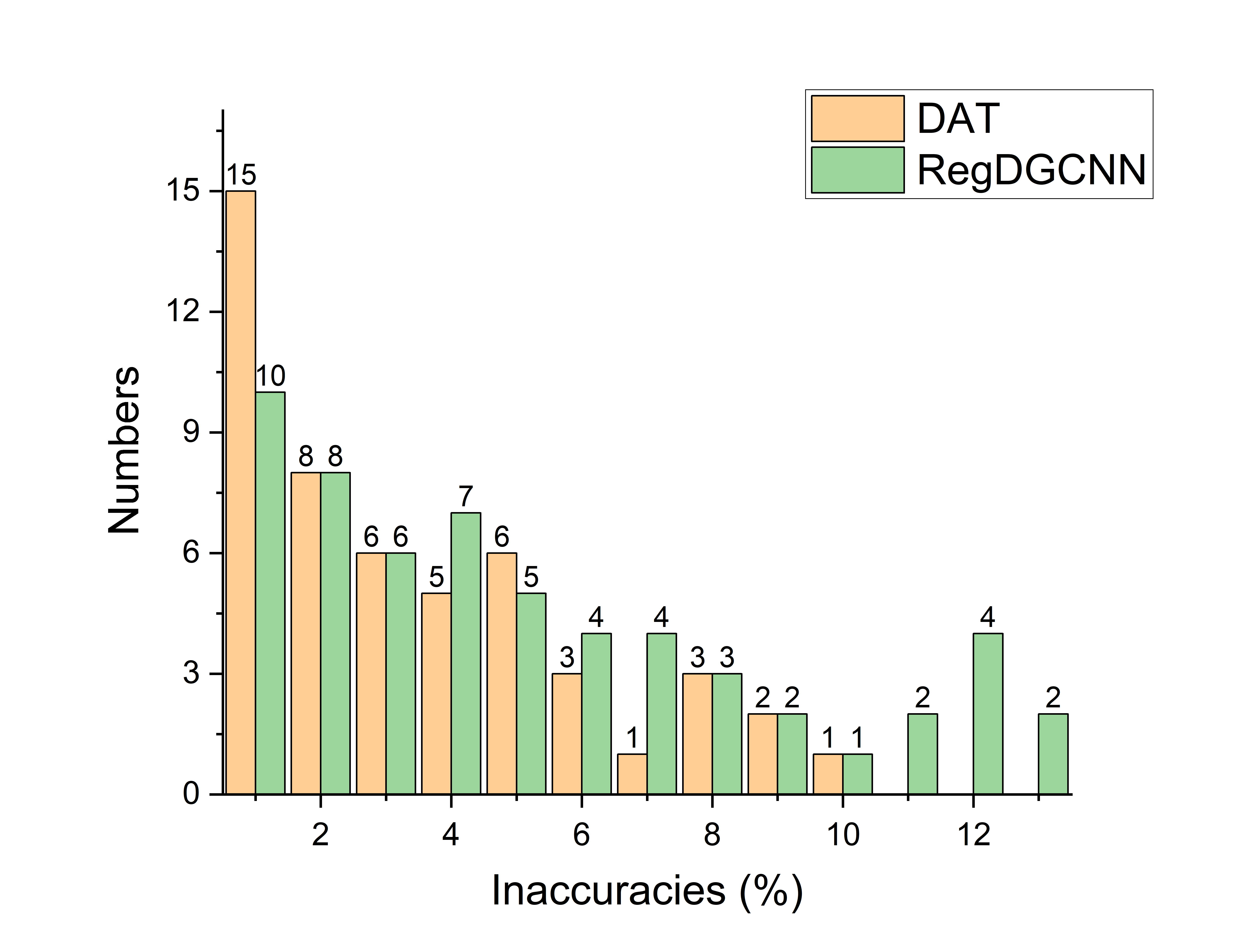}
        \caption{ORV Error Distribution}
    \end{subfigure}
    \hfill
    \begin{subfigure}{0.43\textwidth}
        \centering
        \includegraphics[width=\textwidth]{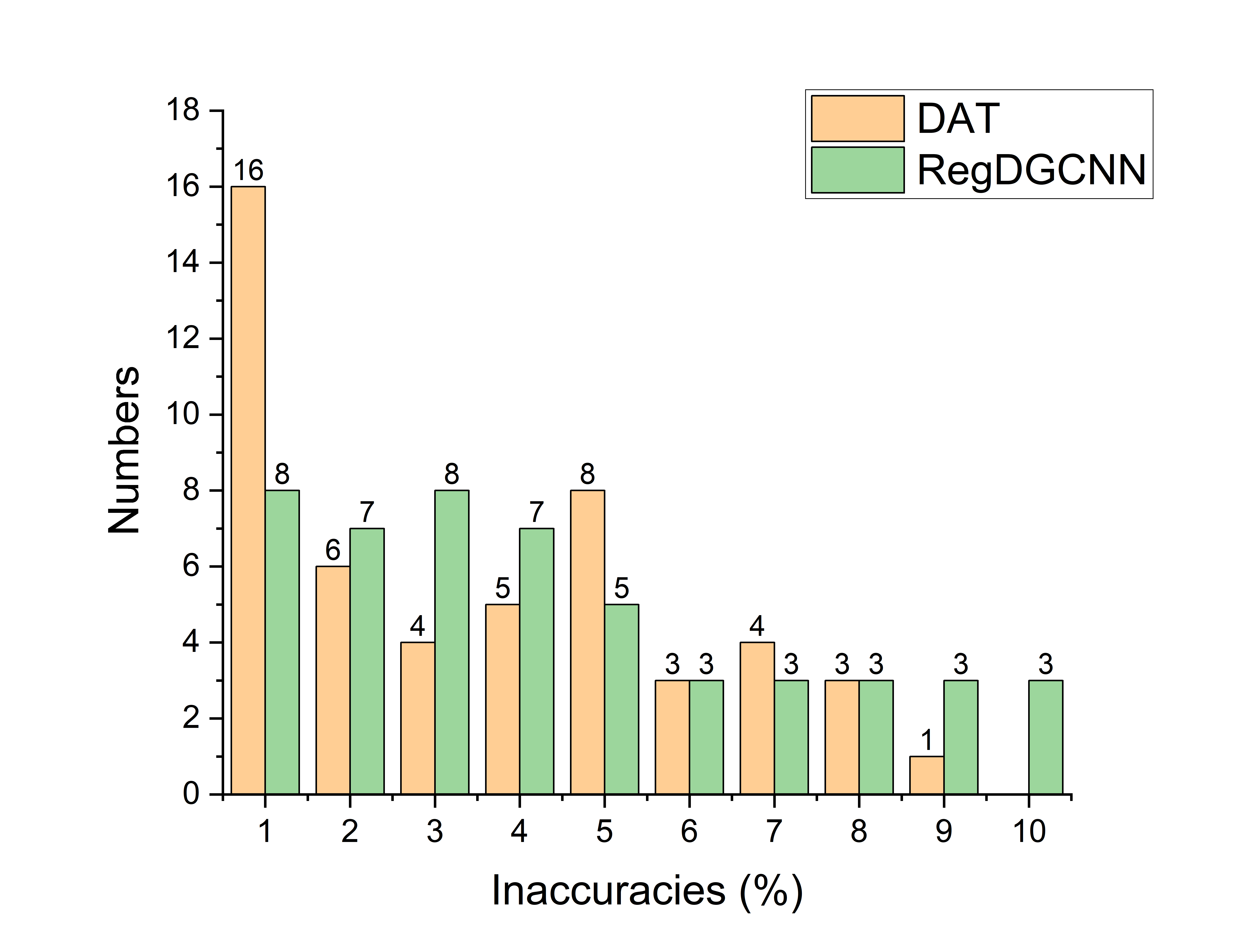}
        \caption{MPV Error Distribution}
    \end{subfigure}
    \caption{Comparison of PCV, SUV, ORV, and MPV Error Distributions}
    \label{fig:4V}
\end{figure*}

\begin{figure}[h]
    \centering
    \includegraphics[width=0.45\textwidth]{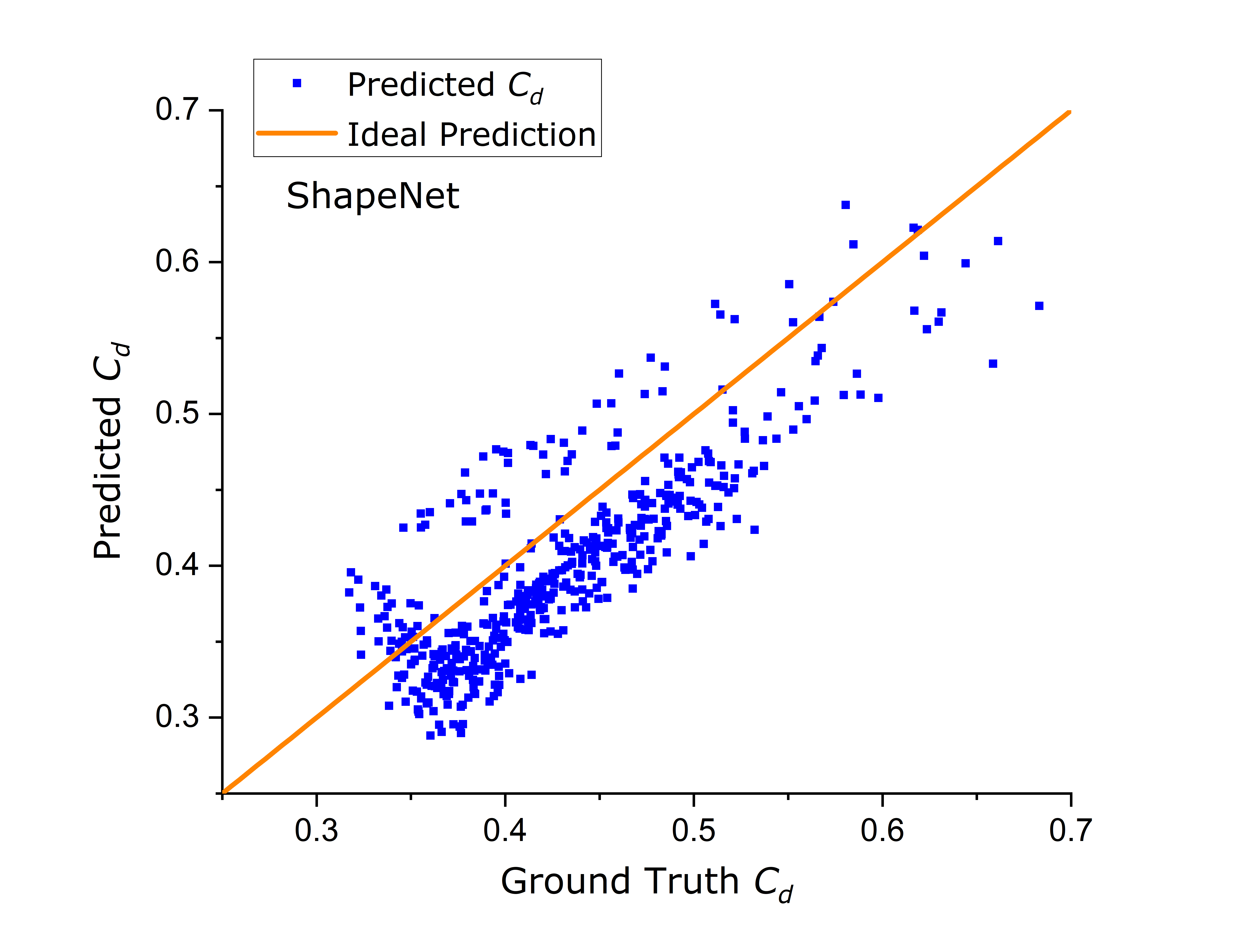}
    \caption{Normal distribution of New Validation Set}
    \label{fig:ShapeNet}
\end{figure}

To further evaluate model generalization, we conducted additional validation using an extended dataset derived from wind tunnel experimental data. The comparative prediction results of DAT and RegDGCNN are shown in Fig.~\ref{fig:new ed}. The results confirm that DAT maintains high accuracy even when applied to previously unseen geometries, demonstrating its strong extrapolation capabilities beyond the training distribution. The complete scatter plot of predicted versus ground truth drag coefficients (\(C_d\)) across the DrivAerNet++ test set is provided in the Supplementary Material, where DAT consistently outperforms RegDGCNN across all subsets.

To investigate performance variation across different vehicle body types, the test samples were categorized into PCV, SUV, ORV, and MPV classes (see Fig.~\ref{fig:4V}). The average prediction error in each category is summarized in Tab.~\ref{tab:example}. As shown in Tab.~\ref{tab:example}, the DAT model consistently achieves lower prediction errors than the RegDGCNN baseline across all vehicle categories. Notably, error rates drop from 4.650\% to 2.553\% for PCV models and from 5.428\% to 3.025\% for MPV models, highlighting DAT’s improved generalization to geometrically diverse and aerodynamically complex vehicle types.

\begin{table}[h]
\caption{\label{tab:example} \(C_d\) and Errors for Different Vehicle Models.}
\begin{ruledtabular}
\begin{tabular}{lccc}
\textbf{Model} & \textbf{\(C_d\)} & \textbf{RegDGCNN} & \textbf{DAT} \\ 
\hline
PCV  & 0.26 - 0.34 & 4.650\% & 2.553\% \\
SUV  & 0.28 - 0.35 & 5.099\% & 3.374\% \\
ORV  & 0.32 - 0.40 & 4.805\% & 3.916\% \\
MPV  & 0.31 - 0.38 & 5.428\% & 3.025\% \\
\end{tabular}
\end{ruledtabular}
\end{table}

The results indicate that while RegDGCNN performs adequately on typical shapes (e.g., PCV), its accuracy declines for categories such as SUV, ORV, and MPV, which involve greater structural variation and nonlinear flow features. This is likely due to the limited receptive field of local graph convolutions, which struggle to capture long-range dependencies and global context.

In contrast, DAT incorporates a self-attention mechanism that enhances the model’s ability to identify key shape features—such as windshield inclination, rear-end taper, and underbody contours—that significantly impact aerodynamic drag. This enables DAT to maintain accuracy even for shapes that deviate from the training distribution.

From an application perspective, DAT achieves prediction errors generally within 2\%–3\%, which meets or exceeds the tolerance required by most automotive manufacturers during early-stage aerodynamic assessment. These results demonstrate DAT's practical viability as a fast and accurate surrogate model for aerodynamic design workflows.

\begin{figure*}[t]
    \centering
    \begin{subfigure}{0.47\textwidth}
        \centering
        \includegraphics[width=\textwidth]{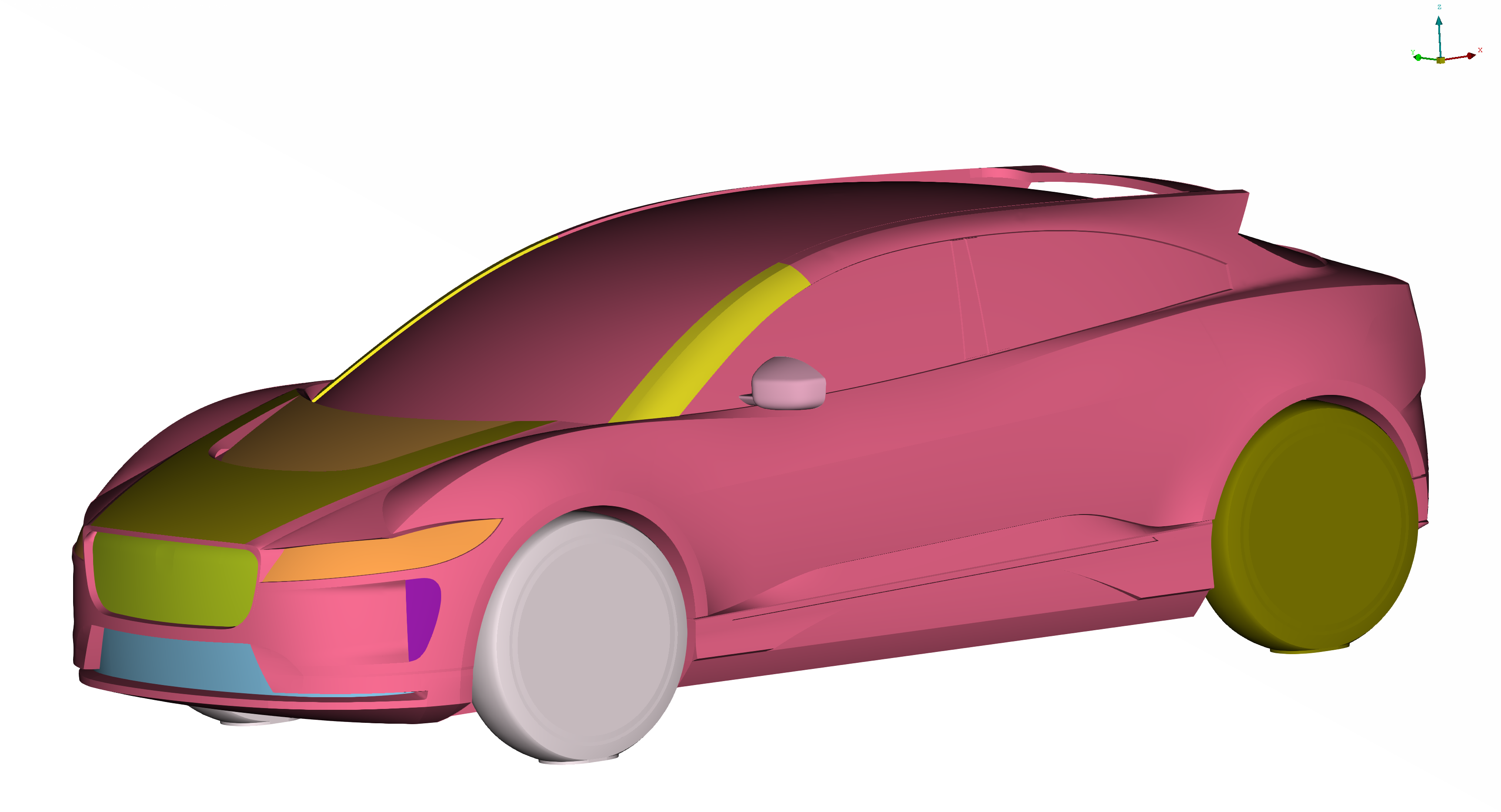}
        \caption{Basic Model}
        \label{fig:new_predict1}
    \end{subfigure}
    \hfill
    \begin{subfigure}{0.47\textwidth}
        \centering
        \includegraphics[width=\textwidth]{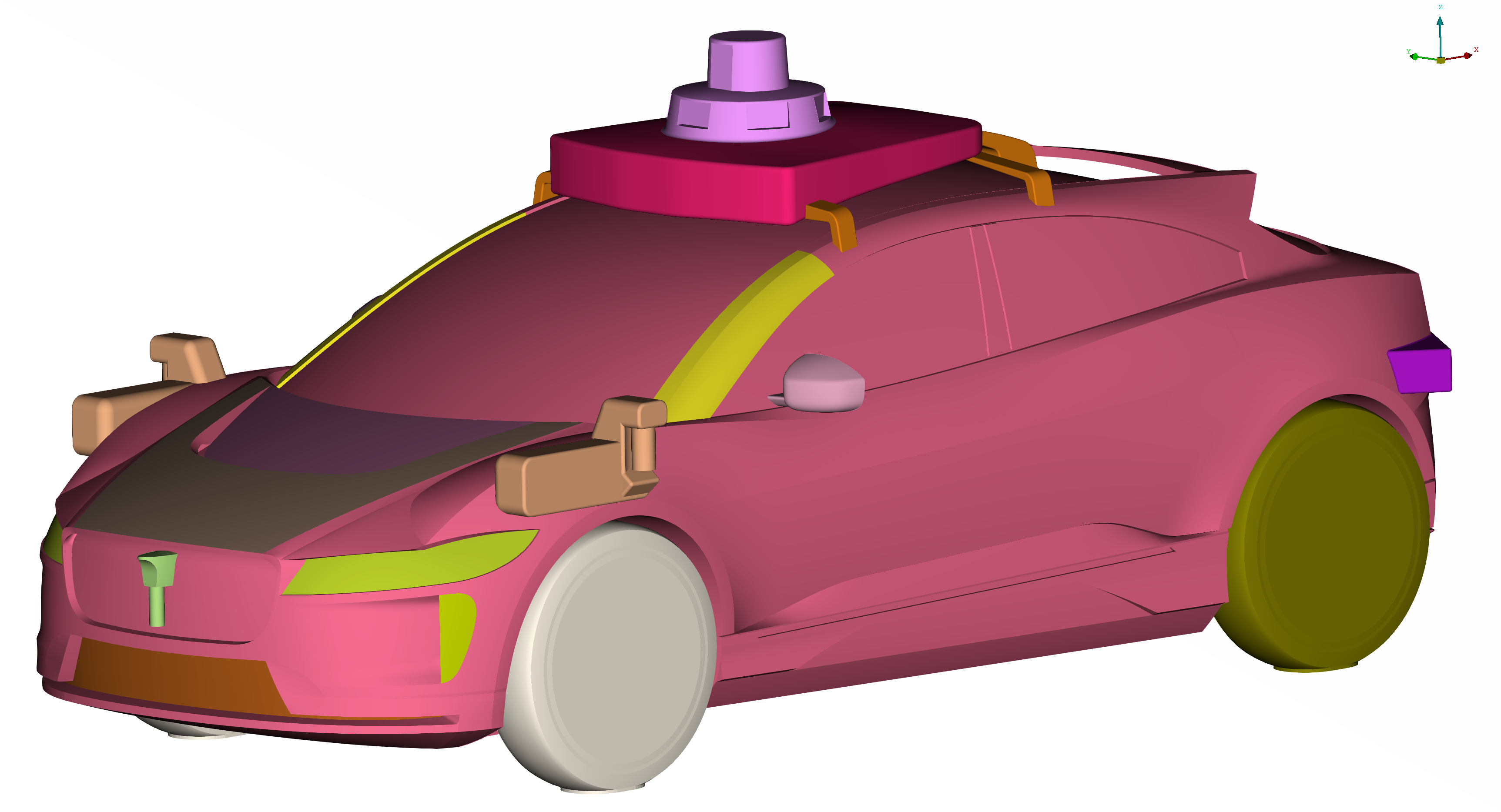}
        \caption{ADAS-Enhanced Model}
        \label{fig:newpredict3}
    \end{subfigure}
    \caption{Error Distribution Across Different Vehicle Types}
    \label{fig:ADAScar}
\end{figure*}

In order to validate the accuracy and applicability of the study, this study provides quantitative results on other publicly available ShapeNet vehicle datasets to demonstrate the robustness and generalization of the proposed method beyond the DrivAerNet dataset\cite{chang2015shapenet, umetani2018learning}. However, the ShapeNet vehicle dataset features an average mesh resolution of approximately 60 to 100 thousand faces, which is lower than the 8–16 million faces typical of DrivAerNet. This difference in geometric fidelity may introduce some bias in the predicted aerodynamic coefficients due to increased surface roughness and reduced detail in key flow regions. As shown in Fig.~\ref{fig:ShapeNet}, the DAT model exhibits a clear tendency of small bias in the prediction results. In this study, it is believed that this phenomenon is due to the low resolution of the ShapeNet model, which leads to an increase in its geometric roughness, thus making the air drag coefficient obtained from CFD calculations to be on the large side, with most of the values centered in the 0.4 - 0.5 range. On the other hand, the drag coefficient in the DrivAerNet dataset is mainly centered around 0.3, which is more in line with the wind resistance data of real cars. In addition, in the validation of the supplementary test set, it can be observed that the DAT model performs better when dealing with the high-resolution vehicle point cloud data. However, the pre-processing (e.g., slicing) and point cloud reading time of the model will be increased accordingly, but the overall prediction time can still be controlled within 10 seconds, and the prediction error is maintained within 6\%.

\label{sec:discussion}
\subsection{Discussion}
After analyzing the two architectures RegDGCNN and DAT from a variety of car model datasets, the research results show that DAT has a better generalization to untrained datasets and the computational accuracy meets the needs of the existing automobile development, and this feature is an important step for the application of machine learning methods in real production. However, our team encountered the following problems when validating a variety of models: 

\begin{enumerate}
    \item As smart driving technology began to be gradually applied to family cars, many models began to be retrofitted with smart driving devices, such as LIDAR, ultrasonic radar and vision camera components on the original model, and a comparison of the retrofitted car model is shown in Fig.~\ref{fig:ADAScar}. Such additional components cannot be well adapted to the change by either RegDGCNN or DAT architectures, and the specific parameters are shown in Tab.~\ref{tab:ADAScar}. It can be clearly seen that the car retrofitted with intelligent driving devices is then predicted with a significant deviation, which is far beyond what is acceptable in the car development process.

\begin{table}[h]
\caption{\label{tab:ADAScar} Error Comparison between Basic and Enhanced Models.}
\begin{ruledtabular}
\begin{tabular}{lcc}
\textbf{Model} & \textbf{DAT} & \textbf{RegDGCNN} \\ 
\hline
Basic Model    & 0.298 (3.84\%)  & 0.284 (8.68\%)  \\
Enhanced Model & 0.304 (10.85\%) & 0.298 (12.61\%) \\
\end{tabular}
\end{ruledtabular}
\end{table}
    \item For some minor structural changes, such as adjusting the angle of the rear spoiler, changing the shape of the mirrors, etc., and adding built-in radiator parts. The machine learning method may not be able to accurately predict the change trend after the change, for example, the change of the angle of the rear spoiler will make the drag coefficient of the whole car produce a change of decreasing and then increasing\cite{cheng2019experimental, das2017cfd}. After comparing the data through deep learning calculations, it is found that such subtle trend changes are likely to be covered by the bias generated by each prediction to the extent that a generalized pattern cannot be derived. This study hypothesizes that this is a phenomenon that may be due to insufficient training samples.
\end{enumerate}

\section{Conclusion}
In this paper, we obtain automotive aerodynamic performance data through high-fidelity CFD numerical simulation, construct a large-scale multimodal dataset DrivAerNet++ containing multiple design parameters and aerodynamic coefficients, and establish a DAT prediction model based on geometric deep learning, and validate the performance differences of different models without using the samples of the test set. The algorithm is innovative in the following two aspects: 1. Utilizing the negative feedback regulation mechanism in the field of automatic control, the idea of error-correcting feedback structure to comprehensively capture the local features of the point cloud; 2. Utilizing the attention module based on the affinity of the channel to help avoid possible redundancy in feature mapping. The conclusions of the study are as follows:

\begin{enumerate}
    \item Utilizing the DrivAerNet++ dataset with the geometric deep learning model, the vehicle drag coefficients can be predicted efficiently, and the model's errors on both the initial and extended validation sets are controlled within 4\%. Compared with RegDGCNN and DGCNN, the DAT model has higher accuracy and generalization ability, and can directly process 3D mesh data without additional image rendering or SDF construction, which improves the efficiency and practicability of pre-processing.
    
    \item The generalization ability of the model is enhanced by expanding the validation set by introducing SUV and MPV models. Compared with the RegDGCNN model, the DAT model has lower prediction error in each drag coefficient \(C_d\) range, and shows better adaptability and complex shape prediction capability. Combined with wind tunnel measurements, DAT not only meets the current industrial accuracy requirements, but also shows the potential to be applied to large-scale multimodal data.
    
    \item Despite the excellent performance of DAT, the study still finds that the training and optimization process has the problems of computational resource constraints, data processing complexity, and high cost of wind tunnel tests. Future research should focus on simplifying data management and loading strategies, improving computational efficiency, and exploring more cost-effective validation methods to promote the application of DAT in engineering practice. Through continuous optimization, DAT is expected to have a wide and far-reaching impact on the prediction of automotive aerodynamic performance and related fields.
\end{enumerate}

\section*{Supplementary Material}
This supplementary document provides an additional visualization supporting the main manuscript. The figure shows the configuration-specific predicted \(C_d\) values across various subsets of the DrivAerNet++ test set.

\section*{Acknowledgment}
This work was supported by the National Natural Science Foundation of China (Grant No. 52472416).

\begin{figure*}[t]
    \centering
    \begin{subfigure}{0.4\textwidth}
        \centering
        \includegraphics[width=\linewidth]{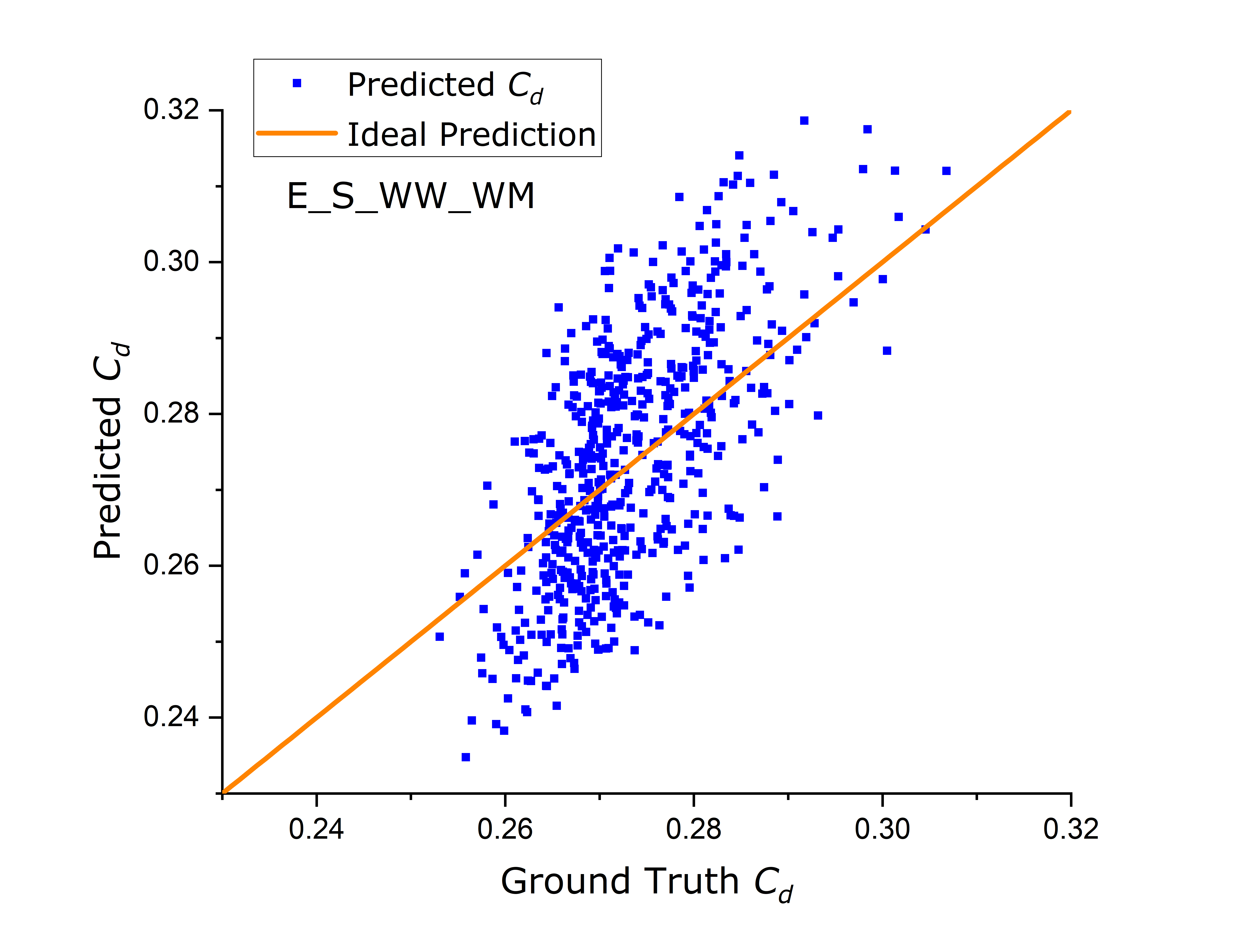}
        \caption{Predicted \(C_d\) on E\_S\_WW\_WM Set}
    \end{subfigure}
    \hfill
    \begin{subfigure}{0.4\textwidth}
        \centering
        \includegraphics[width=\linewidth]{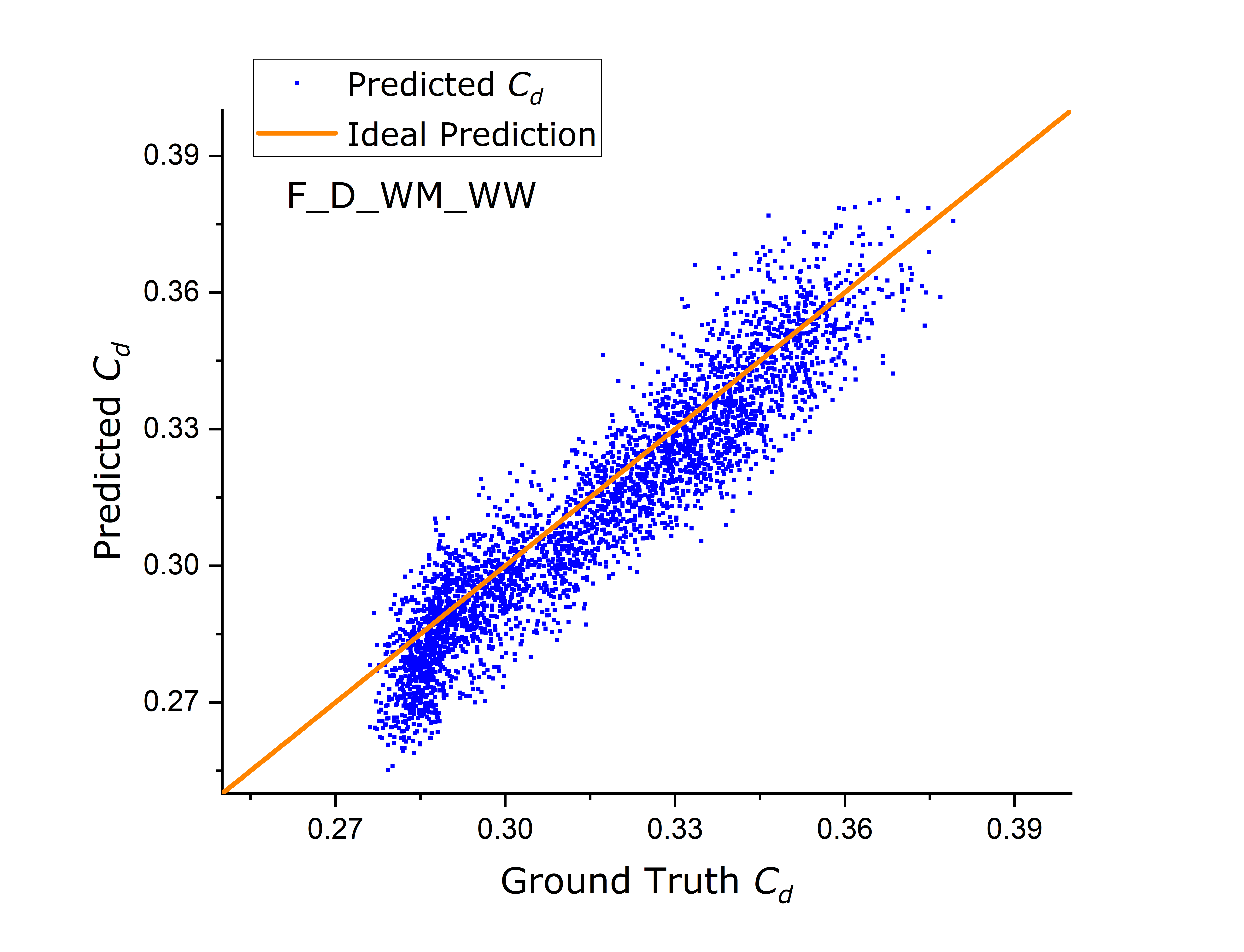}
        \caption{Predicted \(C_d\) on F\_D\_WM\_WW Set}
    \end{subfigure}
    \begin{subfigure}{0.4\textwidth}
        \centering
        \includegraphics[width=\linewidth]{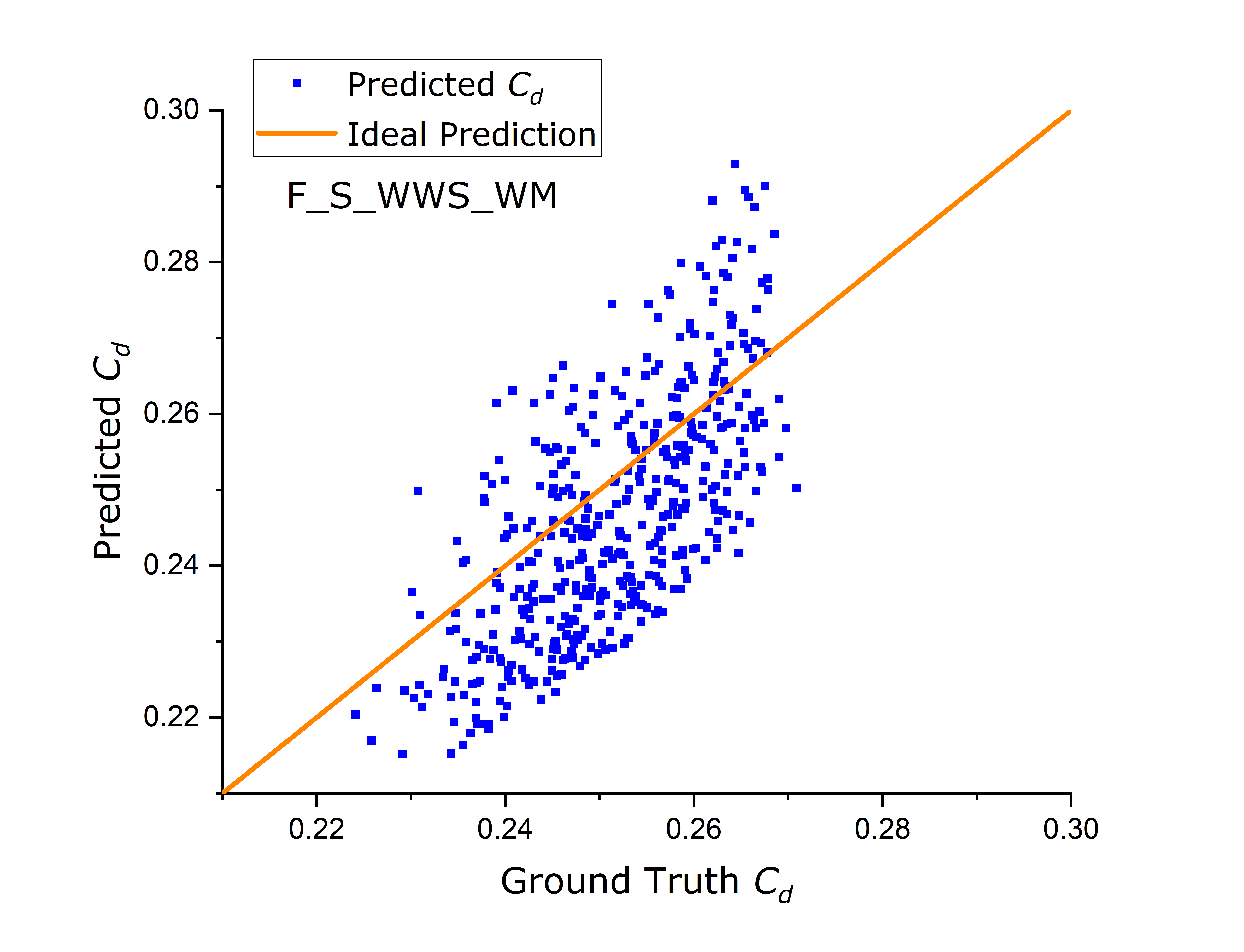}
        \caption{Predicted \(C_d\) on F\_S\_WWS\_WM Set}
    \end{subfigure}
    \hfill
    \begin{subfigure}{0.4\textwidth}
        \centering
        \includegraphics[width=\linewidth]{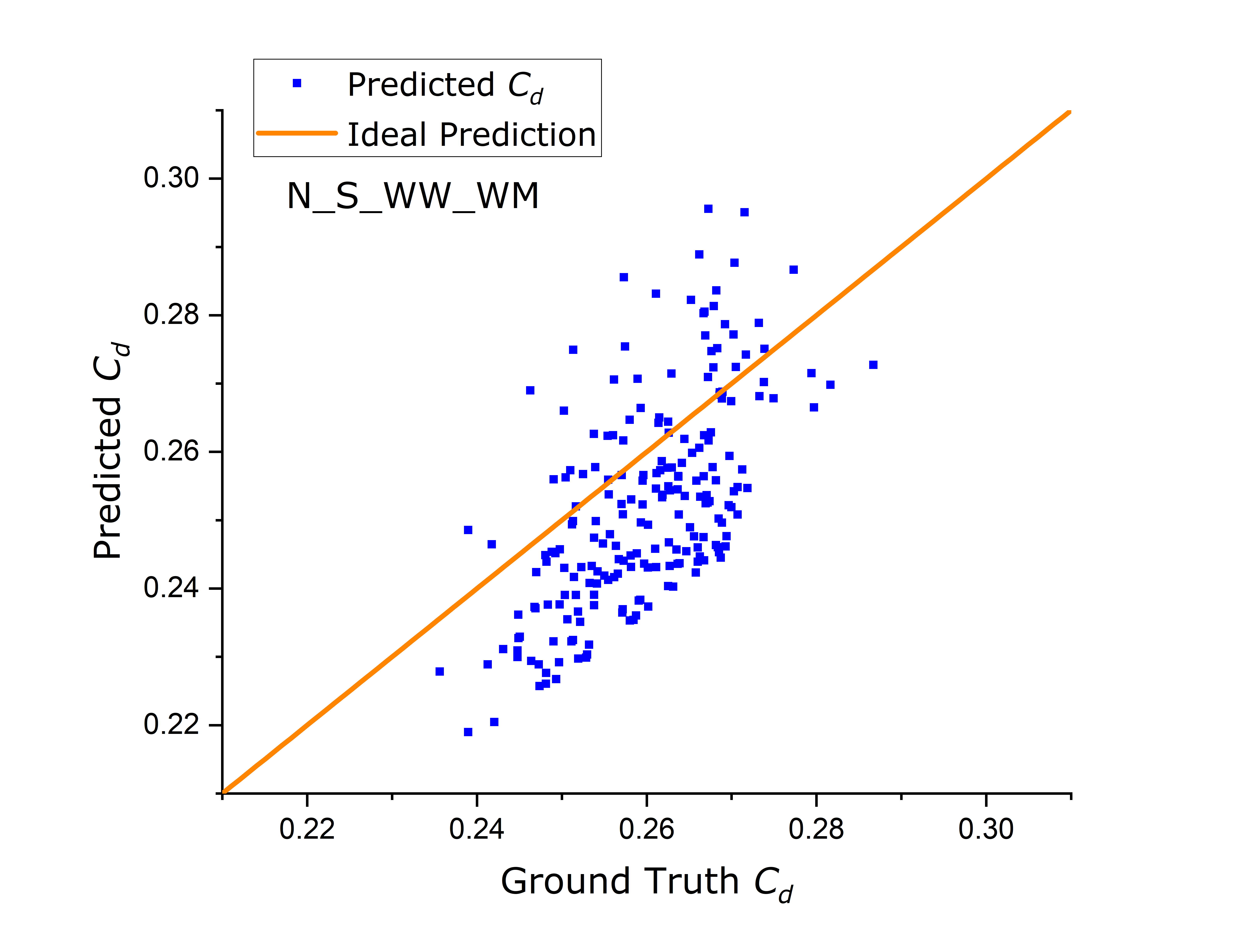}
        \caption{Predicted \(C_d\) on N\_S\_WW\_WM Set}
    \end{subfigure}
    \begin{subfigure}{0.4\textwidth}
        \centering
        \includegraphics[width=\linewidth]{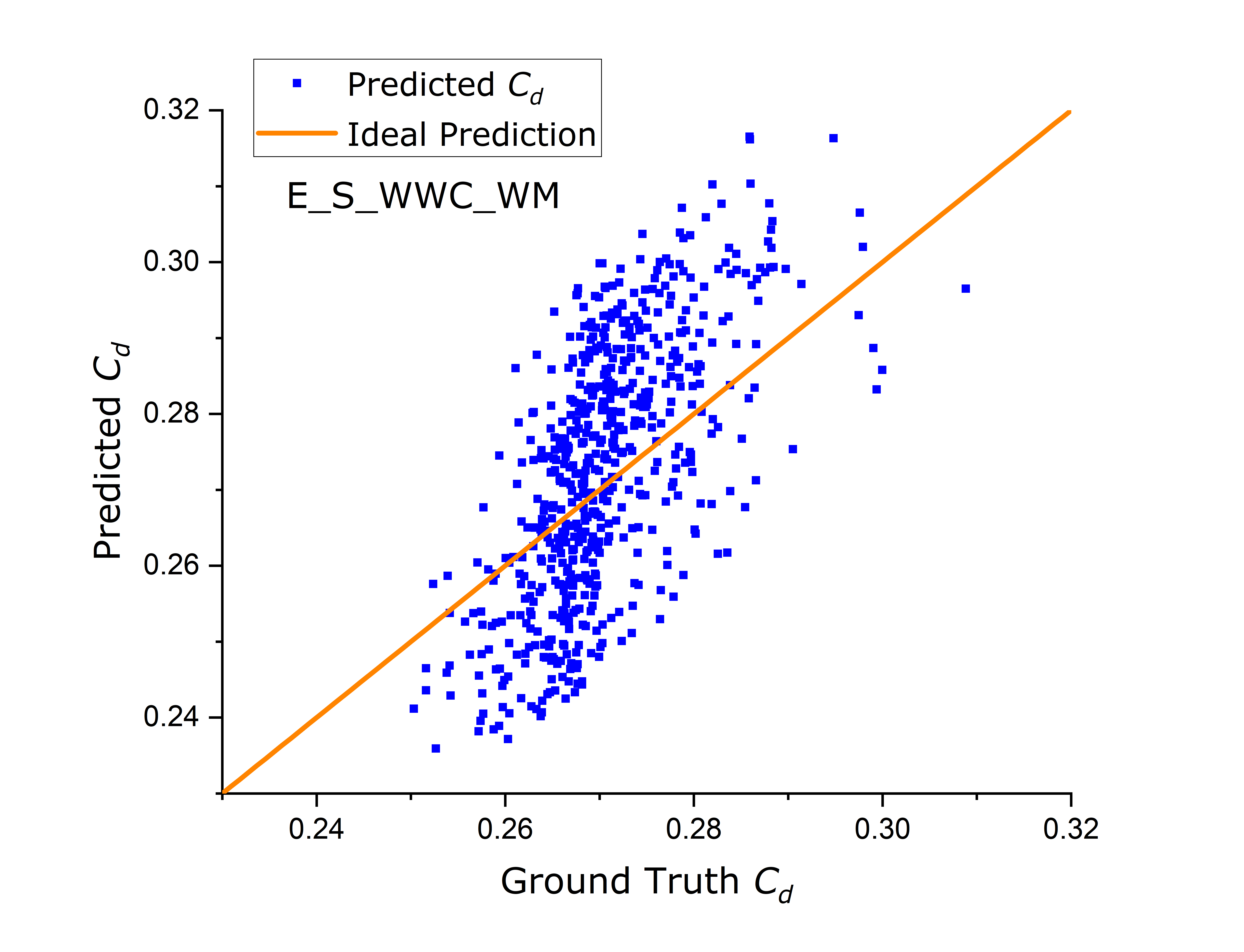}
        \caption{Predicted \(C_d\) on E\_S\_WWC\_WM Set}
    \end{subfigure}
    \hfill
    \begin{subfigure}{0.4\textwidth}
        \centering
        \includegraphics[width=\linewidth]{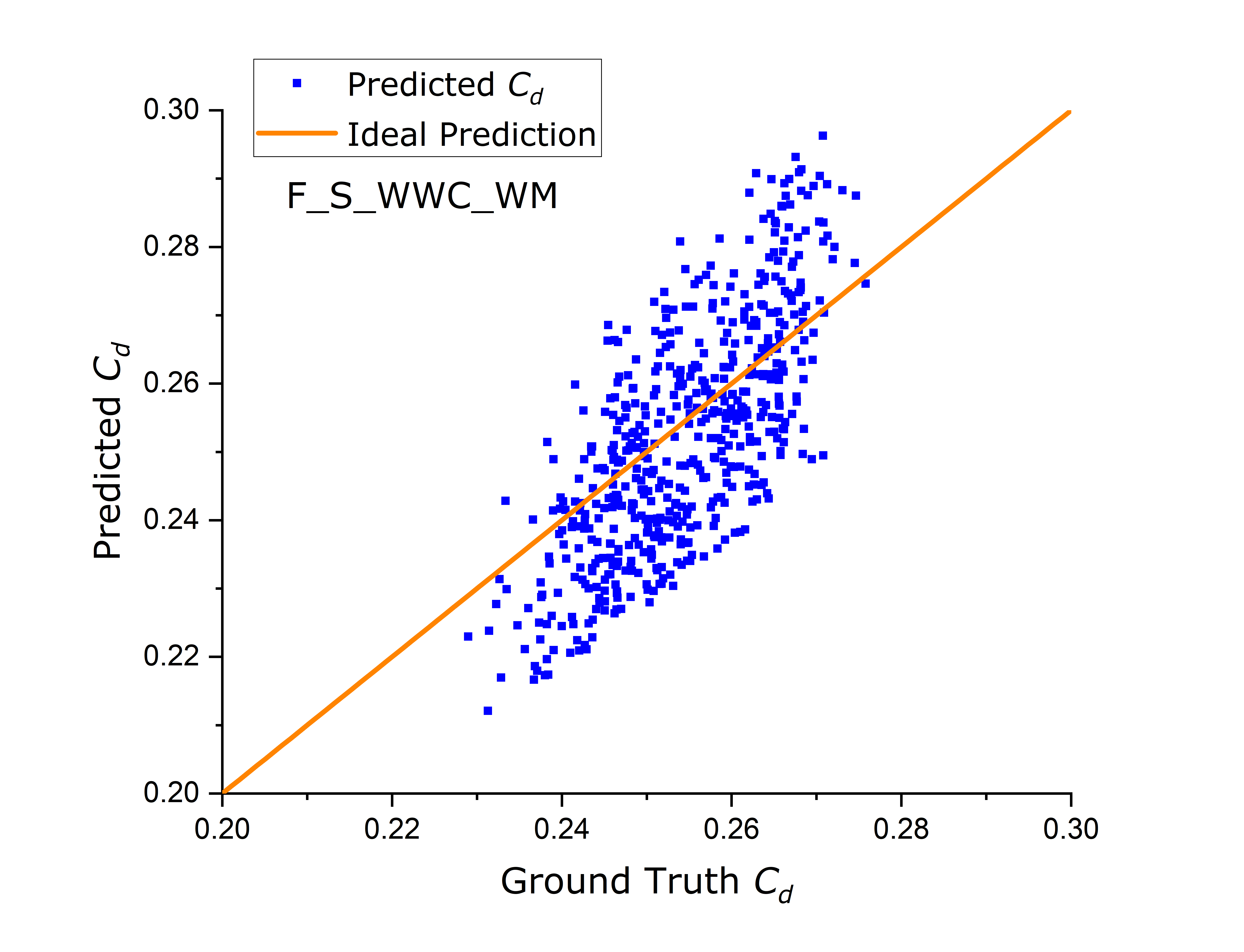}
        \caption{Predicted \(C_d\) on F\_S\_WWC\_WM Set}
    \end{subfigure}
    \begin{subfigure}{0.4\textwidth}
        \centering
        \includegraphics[width=\linewidth]{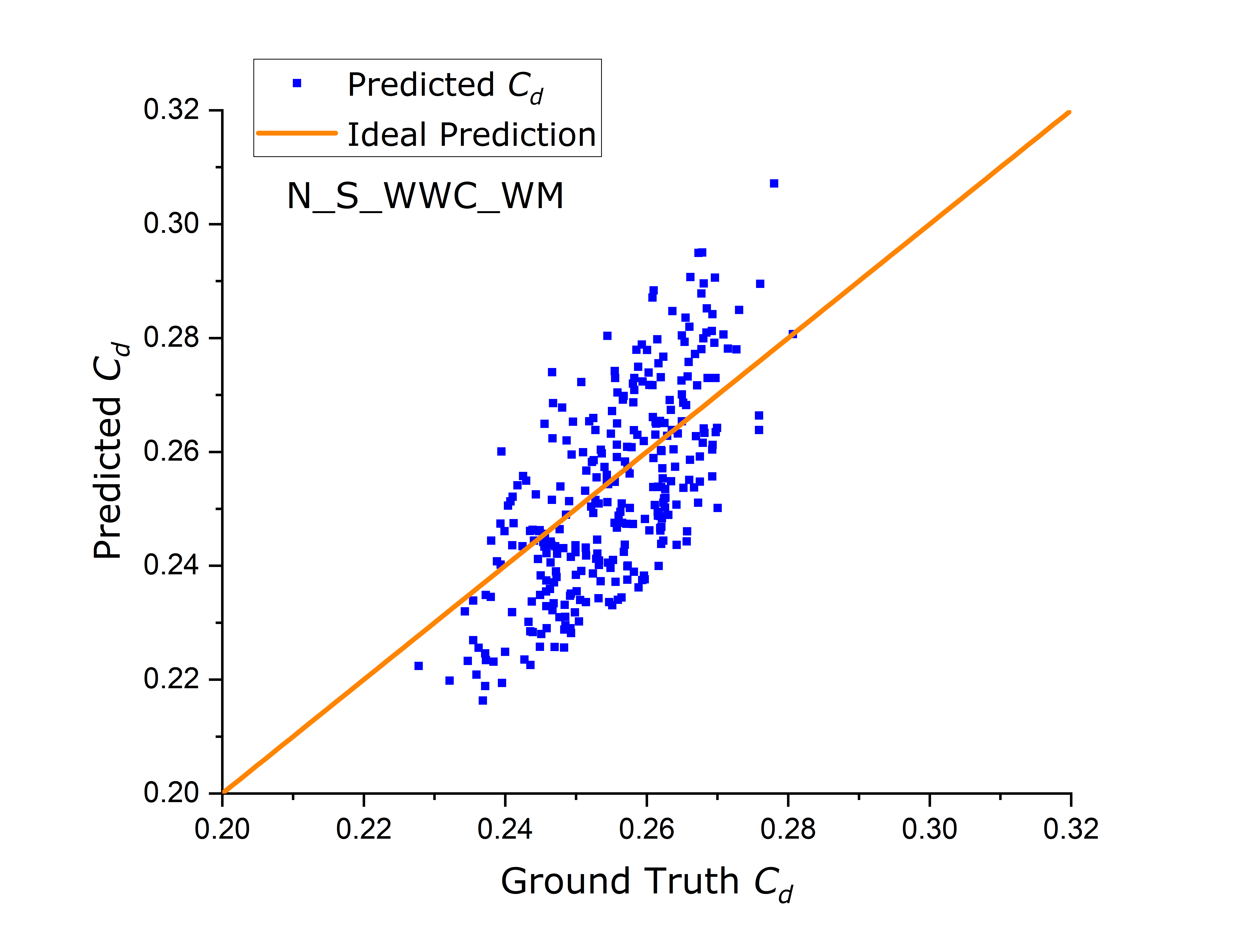}
        \caption{Predicted \(C_d\) on N\_S\_WWC\_WM Set}
    \end{subfigure}
    \hfill
    \begin{subfigure}{0.4\textwidth}
        \centering
        \includegraphics[width=\linewidth]{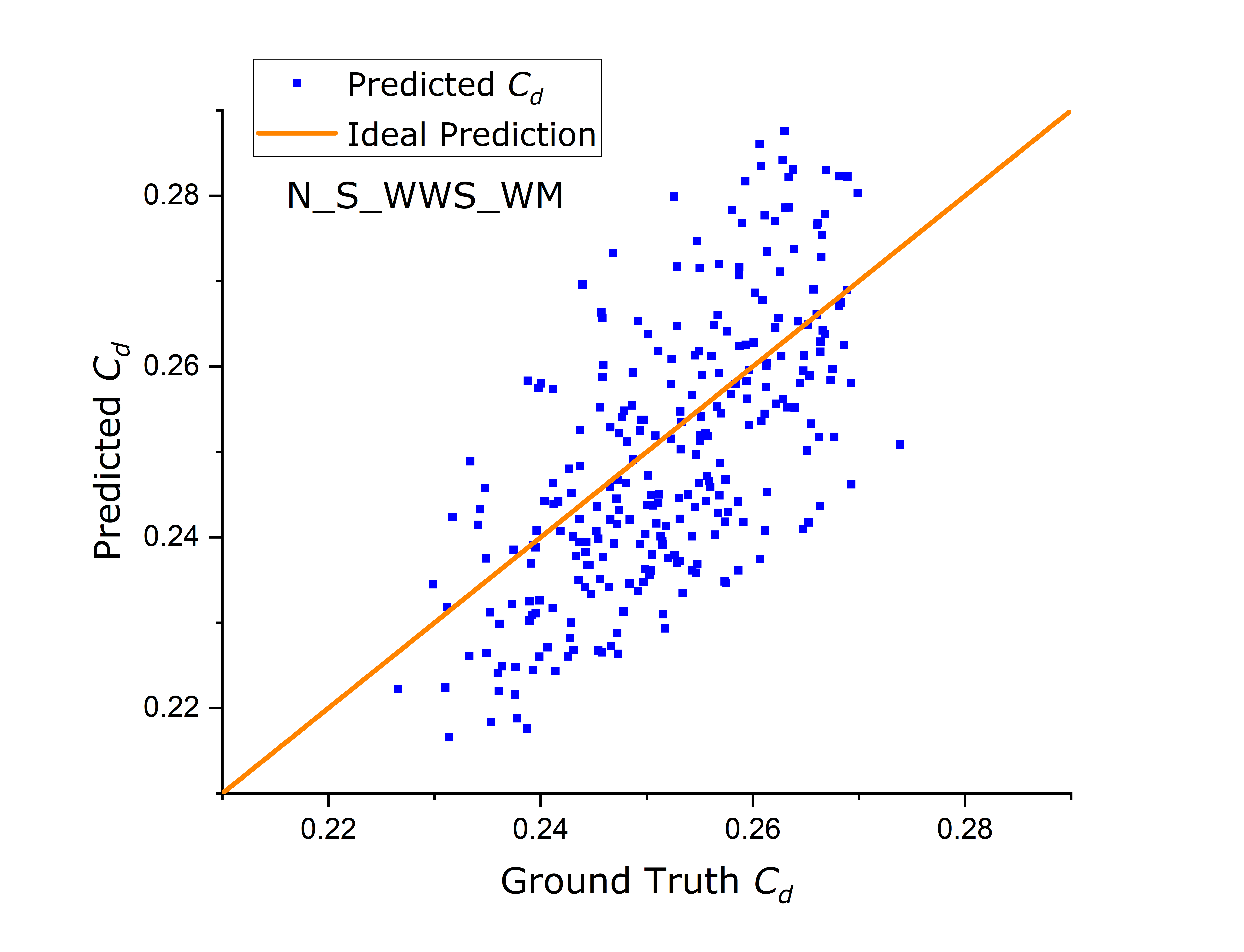}
        \caption{Predicted \(C_d\) on N\_S\_WWS\_WM Set}
    \end{subfigure}
    \caption{Configuration-specific predicted \(C_d\) values across multiple subsets of the DrivAerNet++ test set.}
    \label{fig:Material}
\end{figure*}

\end{document}